\documentclass[%
 reprint,
 amsmath,amssymb,
 aps,
]{revtex4-1}

\usepackage{graphicx}
\usepackage{dcolumn}
\usepackage{bm}
\usepackage{subcaption}
\usepackage{tikz}
\usetikzlibrary{calc,arrows.meta,positioning}
\usepackage{hyperref}

\usepackage{floatrow}
\usepackage[export]{adjustbox}

\usepackage{tabularray}

\begin{document}

\preprint{APS/123-QED}

\title{Parameter estimation from an Ornstein-Uhlenbeck process with measurement noise}
\author{Simon Carter}
 \affiliation{Applied Mathematics and Laufer Center for Physical and Quantitative Biology, Stony Brook University, Stony Brook NY 11794-5281.}
\author{Lilianne R. Mujica-Parodi}
\affiliation{Biomedical Engineering Department and Laufer Center for Physical and Quantitative Biology, Stony Brook University, Stony Brook NY 11794-5281.}
\author{Helmut H. Strey}
\affiliation{Biomedical Engineering Department and Laufer Center for Physical and Quantitative Biology, Stony Brook University, Stony Brook NY 11794-5281.}

\date{\today}

\begin{abstract}

This article aims to investigate the impact of noise on parameter fitting for an Ornstein-Uhlenbeck process, focusing on the effects of multiplicative and thermal noise on the accuracy of signal separation. To address these issues, we propose algorithms and methods that can effectively distinguish between thermal and multiplicative noise and improve the precision of parameter estimation for optimal data analysis. Specifically, we explore the impact of both multiplicative and thermal noise on the obfuscation of the actual signal and propose methods to resolve them. First, we present an algorithm that can effectively separate thermal noise with comparable performance to Hamilton Monte Carlo (HMC) but with significantly improved speed. We then analyze multiplicative noise and demonstrate that HMC is insufficient for isolating thermal and multiplicative noise. However, we show that, with additional knowledge of the ratio between thermal and multiplicative noise, we can accurately distinguish between the two types of noise when provided with a sufficiently large sampling rate or an amplitude of multiplicative noise smaller than thermal noise.
Thus, we demonstrate the mechanism underlying an otherwise counterintuitive phenomenon: when multiplicative noise dominates the noise spectrum, one can successfully estimate the parameters for such systems after adding additional white noise to shift the noise balance.

\end{abstract}

\pacs{Valid PACS appear here}
\maketitle

\onecolumngrid

\subsection{Introduction}
Parameter estimation from data is a critical aspect of statistical modeling and machine learning. It enables the determination of the ideal values for the parameters of a model given observation data, and accurate parameter estimation is essential for obtaining meaningful results from a model \cite{RN42}. Additionally, incorporating prior knowledge and constraints into the estimation process allows for a more robust and interpretable model at the cost of the accuracy of those assumptions \cite{RN90}. Moreover, the estimation process helps to identify patterns and relationships in the data and make predictions based on those patterns \cite{RN42}. 
The Ornstein-Uhlenbeck (OU) process \cite{RN28} is a widely used and important mathematical tool for modeling and understanding various real-world phenomena. It has been widely studied in physics, finance, biology, neuroscience, and engineering. The OU process is a continuous-time Markov process characterized by its mean-reverting behavior towards a long-term average. This makes it well-suited for modeling situations where there is a tendency for a variable to return to a long-term average while also allowing for short-term fluctuations. The OU process has been used in finance to model the dynamics of stock prices and interest rates \cite{RN103}, in biology to model the evolution of population sizes \cite{RN104}, in neuroscience to model neural activity \cite{Maller2009, Ricciardi1979-bg}, and in engineering to model the behavior of control systems \cite{RN105}. Previously, we had developed an analytical solution for the maximum likelihood parameter fitting of an Ornstein-Uhlenbeck process without measurement noise \cite{RN91}.
Measurement noise is a ubiquitous issue in parameter estimation from data, and it can significantly impact the accuracy and reliability of the estimated parameters. Measurement noise refers to the inaccuracies and fluctuations in the measurement process that result in data points that deviate from the true underlying relationship \cite{RN102}. The presence of measurement noise can obscure the true patterns and relationships in the data, making it more challenging to estimate the parameters of a model accurately \cite{RN42}. Moreover, measurement noise can also increase the variability of the data, leading to increased uncertainty in the estimated parameters. This can result in less precise and unreliable estimates and increase the risk of overfitting the data, where the model fits the measurement noise rather than the true underlying relationship.

The challenge of noise removal has been explored in many different aspects traditionally managed through filtering or averaging. Here, we present a Bayesian approach that models the data and the noise probabilistically and simultaneously. Specifically, we look at two types of noise: additive Gaussian (white) and dependant (multiplicative) noise. White noise is the most commonly modeled and assumed noise. However, signal-dependent noise can be the dominant source of noise in many cases, with the most prominent examples found in computer vision and neuronal models. An approximation of this signal-dependent noise as white noise may not be sufficient. \cite{Hasinoff2014, LANSKY2001132}

In this paper, we introduce methods for parameter fitting of an Ornstein-Uhlenbeck (OU) process with added white noise and a combination of white noise and multiplicative noise. For the case of white noise alone, we developed an expectation-maximization algorithm that estimates the parameters from data. Our method returns parameter values similar to those obtained using more general Hamilton Monte Carlo (HMC) methods but are computationally more efficient. When multiplicative noise is added to an OU process (signal), even HMC methods cannot adequately separate the noise from the signal. This is likely due to the similarity between the power spectra of the multiplicative noise and the signal. We found that the OU signal parameters, as well as the amplitudes of the white noise and multiplicative noise, can be estimated using probabilistic modeling if the ratio of multiplicative to white noise does not exceed a threshold that depends on the sampling rate of the data. We concluded that to successfully model a mixture of white and multiplicative noise, the ratio of white to multiplicative noise must be known. 

\subsection{Probabilistic Description of Processes}\label{process}
A probabilistic description of an overdamped Brownian particle in a harmonic potential (also called the Ornstein-Uhlenbeck process) was first reported by Ornstein and Uhlenbeck in 1930 \cite{RN28}
\begin{equation}\label{OUp}
	x_{t+\Delta t} \sim \mathcal{N}(\mu=Bx_{t},\sigma^{2}=A(1-B^{2}))
\end{equation}
where $B(\Delta t) = \exp \left( { - \frac{\Delta t}{\tau}} \right)$ with $\tau$ as the relaxation time, and $\mathcal{N}$ representing a normal distribution.
This equation describes the conditional probability of finding a particle at time $t+\Delta t$ at $x_{t+\Delta t}$ given that it was at $x_{t}$ at time $t$.  The likelihood function for a specific time trace that is taken at intervals $\Delta t$, $\left\{x_i(i\Delta t)\right\}$ is:
\begin{multline}
	p\left( \left\{x_i(t_i)\right\} \left| B, A \right.\right) =
	\frac{1}{\sqrt {2 \pi A} }
	\exp \left( { - \frac{{x_1}^2}{2A}}\right)
	\frac{1}{{\sqrt {2\pi A(1-B^{2}(\Delta t))}^{(N-1)} }}
	 \times \exp \left( { - \sum\limits_{i=1}^{N-1}\frac{{{{\left( {x_{i+1} - {x_i}B(\Delta t)} \right)}^2}}}{{2A(1-B^{2}(\Delta t))}}} \right)
\end{multline}
We recently published an analytical solution to the Ornstein-Uhlenbeck maximum likelihood problem \cite{RN91}, and here we want to consider the same problem but with added noise.  Here we will consider Gaussian white noise with variance $\sigma_{N}^2$ and multiplicative noise $\sigma_{M}^{2}x^{2}(t)$ that is added to each data point.  The two noise sources can be described as follows:
\begin{equation}
\label{eqn:noise_label}
	y_{i} \sim \mathcal{N}(\mu=x_{i},\sigma^2=\sigma_{N}^{2}+\sigma_{M}^{2}x_{i}^{2})
\end{equation}
with likelihood function for the measured data $\{y_{i}\}$ given $\{x_{i}\}$ and $\sigma_{N}$:
\begin{equation}
	p\left( \left\{y_i(t_i)\right\} \left| \left\{x_i(t_i)\right\},\sigma_{N} \right.\right) =
	\frac{1}{{\sqrt {2\pi \sigma_{N}^{2}}^{N} }}
	\exp \left( { - \sum\limits_{i=1}^{N}\frac{{{{\left( {y_{i} - x_{i}B(\Delta t)} \right)}^2}}}{{2(\sigma_{N}^{2}+\sigma_{M}^{2}x_{i}^{2})}}} \right)
\end{equation}

Since we are measuring $\{y_{i}\}$, $\{x_{i}\}$ are latent variables.  If we knew $\{x_{i}\}$ and $\{y_{i}\}$ the likelihood function would simply be.
\begin{equation}
	p\left( \{x_{i}\},\{y_{i}\}|A,B,\sigma_{N}\right) = 
	p\left( \left\{y_i(t_i)\right\} \left| \left\{x_i(t_i)\right\},\sigma_{N} \right.\right)
	p\left( \left\{x_i(t_i)\right\} \left| B, A \right.\right)
\end{equation}

\begin{figure}[H]
\begin{center}
\tikzset{
   every node/.style={font=\sffamily\small},
   main node/.style={thick,circle,draw,font=\sffamily\Large}
}
\begin{tikzpicture}[->,>={Stealth[round,sep]},shorten >=1pt,auto,node distance=2.0cm,minimum size=1.5cm]
\node[main node] (1) {$x_{1}$};
\node[main node] (2) [right of=1] {$x_{2}$};
\node[main node] (4) [right of=2] {$x_{n-1}$};
\node[main node] (5) [right of=4] {$x_{n}$};
\node[main node] (6) [right of=5] {$x_{n+1}$};
\node[main node] (8) [right of=6] {$x_{N-1}$};
\node[main node] (9) [right of=8] {$x_{N}$};
\node at ($(2)!.5!(4)$) {\ldots};
\node at ($(6)!.5!(8)$) {\ldots};

\node[main node] (10) [below of=1] {$y_{1}$};
\node[main node] (11) [below of=2] {$y_{2}$};
\node[main node] (12) [below of=4] {$y_{n-1}$};
\node[main node] (13) [below of=5] {$y_{n}$};
\node[main node] (14) [below of=6] {$y_{n-1}$};
\node[main node] (15) [below of=8] {$y_{N-1}$};
\node[main node] (16) [below of=9] {$y_{N}$};

\draw [->] (1) -- (2);
\draw [->] (4) -- (5);
\draw [->] (5) -- (6);
\draw [->] (8) -- (9);

\draw [->] (1) -- (10);
\draw [->] (2) -- (11);
\draw [->] (4) -- (12);
\draw [->] (5) -- (13);
\draw [->] (6) -- (14);
\draw [->] (8) -- (15);
\draw [->] (9) -- (16);
\end{tikzpicture}
\caption{Graph representation of a chain of N values $\{x_{i}\}$ that originate from an OU process with corresponding $\{y_{i}\}$ that represent the measured values containing noise}\label{fig:HMCnodesfig}
\end{center}
\end{figure}

But since we don't observe $\{x_{i}\}$, we need to marginalize over $\{x_{i}\}$:
\begin{equation}
p\left( \{y_{i}\}|A,B,\sigma_{N},\sigma_{M}\right) = \idotsint p\left( \{x_{i}\},\{y_{i}\}|A,B,\sigma_{N},\sigma_{M}\right) \,dx_1 \dots dx_N
\end{equation}
Using Bayes Theorem, we can express the conditional probability of the parameters $\{A,B,\sigma_{N},\sigma_{M}\}$ given the measured data $\{y_{i}\}$.
\begin{equation}
	p\left( \{A,B,\sigma_{N},\sigma_{M}\}|\{y_{i}\},\right) \propto
	p\left( \{y_{i}\}|A,B,\sigma_{N}\right)p\left(A,B,\sigma_{N},\sigma_{M}\right)
\end{equation}
In principle, from this equation, we could estimate the parameters from a measurement of $\{y_{i}\}$ using a maximum likelihood approach.

\section{Parameter Estimation with Added Thermal Noise}\label{EM_alg}
As we can see from the previous equation, a simple maximum likelihood approach is computationally challenging.  Because we are dealing with Gaussian distributions, a typical approach is to take the logarithm of the likelihood and hope that the likelihood factors nicely.  But in our case, because of the marginalization of the hidden variables $\{x_{i}\}$, we are taking the logarithm of integrals over $\{x_{i}\}$ which does not simplify the equation.  On the other hand, we recognize that $p\left( \{x_{i}\},\{y_{i}\}|A,B,\sigma_{N}\right)$ factors nicely.  The Expectation Maximization (EM) algorithm takes advantage of this fact (see \cite{RN90} chapter 9).  The EM algorithm works as follows: (1) pick a starting point for the parameters $\Theta^{old} = \{A^{old},B^{old},\sigma_{N}^{old}\}$; (2) calculate the probability distribution for each hidden variable $\{x_{i}\}$ given these parameters; (3) maximize the following function concerning $\Theta$ to find the new parameters.
\begin{equation}\label{eq:EM_max}
	\mathcal{Q}(\Theta,\Theta^{old}) = \idotsint p\left( \{x_{i}\}|\{y_{i}\},\Theta^{old}\right)\ln p\left( \{x_{i}\},\{y_{i}\}|\Theta\right)\,dx_1 \dots dx_N
\end{equation}
(4) repeat by using $\Theta$ as $\Theta_{old}$ until convergence.  The EM algorithm converges but may find local maxima in the posterior distribution.  Therefore, it is important to vary the starting point for $\Theta$.  To evaluate the integral in eq \ref{eq:EM_max}, we need to calculate $p\left( x_{n}|\{y_{i}\},\Theta^{old}\right)$.  We can again use Bayes rule and the conditional Independence rules of a linear hidden Markov chain:
\begin{equation}\label{margxn}
	\begin{aligned}
	p\left( x_{n}|\{y_{i}\},\Theta^{old}\right)&=\frac{p\left( \{y_{i}\}|x_{n},\Theta^{old}\right)p(x_{n})}{p(\{y_{i}\}|\Theta^{old})}\\
	&=\frac{p(y_{1},\dots,y_{n}|x_{n},\Theta^{old})p(y_{n+1},\dots,y_{N}|x_{n},\Theta^{old})p(x_{n})}{p(\{y_{i}\}|\Theta^{old})}\\
	&=\frac{p(y_{1},\dots,y_{n},x_{n}|\Theta^{old})p(y_{n+1},\dots,y_{N}|x_{n},\Theta^{old})}{p(\{y_{i}\}|\Theta^{old})} \\
	 &=\frac{\alpha(x_{n})\beta(x_{n})}{p(\{y_{i}\}|\Theta^{old})} &
	\end{aligned}
\end{equation}
using the following definition
\begin{equation}
	\begin{aligned}
	\alpha(x_{n})&=p(y_{1},\dots,y_{n},x_{n}|\Theta^{old})\\
	\beta(x_{n})&=p(y_{n+1},\dots,y_{N}|x_{n},\Theta^{old})
	\end{aligned}
\end{equation}
here, we took advantage of the properties of the Markov chain $\{x_{i}\}$.  If we know a specific $x_{n}$, then $x_{n-1}$ and $x_{n+1}$ are independent because the path between the two is blocked (See for example \cite{RN90}).  As a consequence of that, the $y_{i}$ with $i\leq n$ are independent of $y_{i}$ with $i>n$ conditioned on $x_{n}$.  Both $\alpha(x_{n})$ and $\beta(x_{n})$ can be defined and calculated recursively:
\begin{equation}
	\begin{aligned}
	\alpha(x_{n})&=p(y_{n}|x_{n},\Theta^{old})\int \alpha(x_{n-1})p(x_{n}|x_{n-1},\Theta^{old})dx_{n-1}\\
	\beta(x_{n})&=\int \beta(x_{n+1})p(y_{n+1}|x_{n+1},\Theta^{old})p(x_{n+1}|x_{n},\Theta^{old})dx_{n+1}
	\end{aligned}
\end{equation}
with the starting value of
\begin{equation}
	\begin{aligned}
\alpha(x_{1})&\propto \exp \left( { - \frac{{x_1}^2}{2A}}\right)\exp \left( -\frac{{{{\left( {y_{1} - x_{1}} \right)}^2}}}{{2\sigma_{N}^{2}}} \right)\\
\beta(x_{N})&=1
	\end{aligned}
\end{equation}
Here, we have to notice that all probability distributions are Gaussian, which means that the product of two Gaussians, as well as the convolution of two Gaussians, is also a Gaussian.  As a consequence, all $\alpha(x_{n})$ and $\beta(x_{n})$ are Gaussian (except for $\beta(x_{N})$).  For a computation, all we need to calculate is the mean and standard deviation of each $\alpha(x_{n})$ and $\beta(x_{n})$ since the normalization of Gaussians is known.  Similarly, we can calculate the probability distributions of consecutive $x_{n-1}$ and $x_{n}$:
\begin{equation}\label{margxnxnmone}
	\begin{aligned}
	p\left( x_{n-1},x_{n}|\{y_{i}\},\Theta^{old}\right)&=\frac{p\left( \{y_{i}\}|x_{n-1},x_{n},\Theta^{old}\right)p(x_{n-1},x_{n})}{p(\{y_{i}\}|\Theta^{old})}\\
	&=\frac{p(y_{1},\dots,y_{n}|x_{n},\Theta^{old})p(y_{n+1},\dots,y_{N}|x_{n},\Theta^{old})p(x_{n})}{p(\{y_{i}\}|\Theta^{old})}\\
	&=\frac{\alpha(x_{n-1})p(y_{n}|x_{n},\Theta^{old})p(x_{n}|x_{n-1},\Theta^{old})\beta(x_{n})}{p(\{y_{i}\}|\Theta^{old})}
	\end{aligned}
\end{equation}
We can now move to the implementation of the EM algorithm for our specific case: (1) We pick starting values for the parameters ${A,B,\sigma_N}$; (2) We calculate all $p\left( x_{n}|\{y_{i}\},\Theta^{old}\right)$ and $p\left( x_{n-1},x_{n}|\{y_{i}\},\Theta^{old}\right)$ which represent the E step; (3) We calculate the improved parameters by using the procedure for estimating parameters from an OU process without additional measurement noise \cite{RN91}, which yields improved parameters for $A$ and $B$.  For this analysis we only require the expectation values of $x_{1}^{2},x_{N}^2$, the sum of $x_{i}^2$, and $x_{i}x_{i+1}$ which can be calculated using eq.\ref{margxn} and eq.\ref{margxnxnmone}.  The updated value for $\sigma_{N}$ can be obtained by considering the following:
\begin{equation}
	\ln \prod_{i=1}^{N}\frac{1}{\sqrt{2\pi\sigma_{N}^{2}}}\exp\left(-\frac{(x_{i}-y_{i})^2}{2\sigma_{N}^{2}}\right)=-N\ln\sigma_{N}-\frac{1}{2\sigma_{N}^2}\sum_{i=1}^{N}(x_{i}-y_{i})^2
\end{equation}
The maximum of this expression can be found by setting its derivative with respect to $\sigma_N$ to zero, which results in:
\begin{equation}
	\sigma_{N, max}^{2} = \frac{1}{N}\sum_{i=1}^{N}\sigma_{x_{i}}^{2}+\mu_{x_{i}}^{2}+y_{i}^{2}-2\mu_{x_{i}}y_{i}
\end{equation}
where we use the $\sigma_{x_{i}}$ and $\mu_{x_{i}}$ from $p\left( x_{n}|\{y_{i}\},\Theta^{old}\right)=\mathcal{N}(x_{i},\mu=\mu_{x_{i}},\sigma=\sigma_{x_{i}})$.\\
To determine the error of the estimated parameters after convergence of the EM algorithm, we can take advantage of the fact that when the EM algorithm has converged, then $\mathcal{Q}(\Theta,\Theta^{old})$ is equal to the posterior distribution \cite{RN90} and we can use the error estimates for $A$ and $B$ from \cite{RN91}. Since $\sigma_{N}$ is independent from $A$ and $B$ in $\mathcal{Q}(\Theta,\Theta^{old})$, we can write:
\begin{equation}
	d\sigma_{N,max} = \frac{\sigma_{N, max}}{\sqrt{2N}}
\end{equation}

\begin{figure}[H]
\begin{center}
\resizebox{.9\linewidth}{!}{%
\includegraphics[height=4cm]{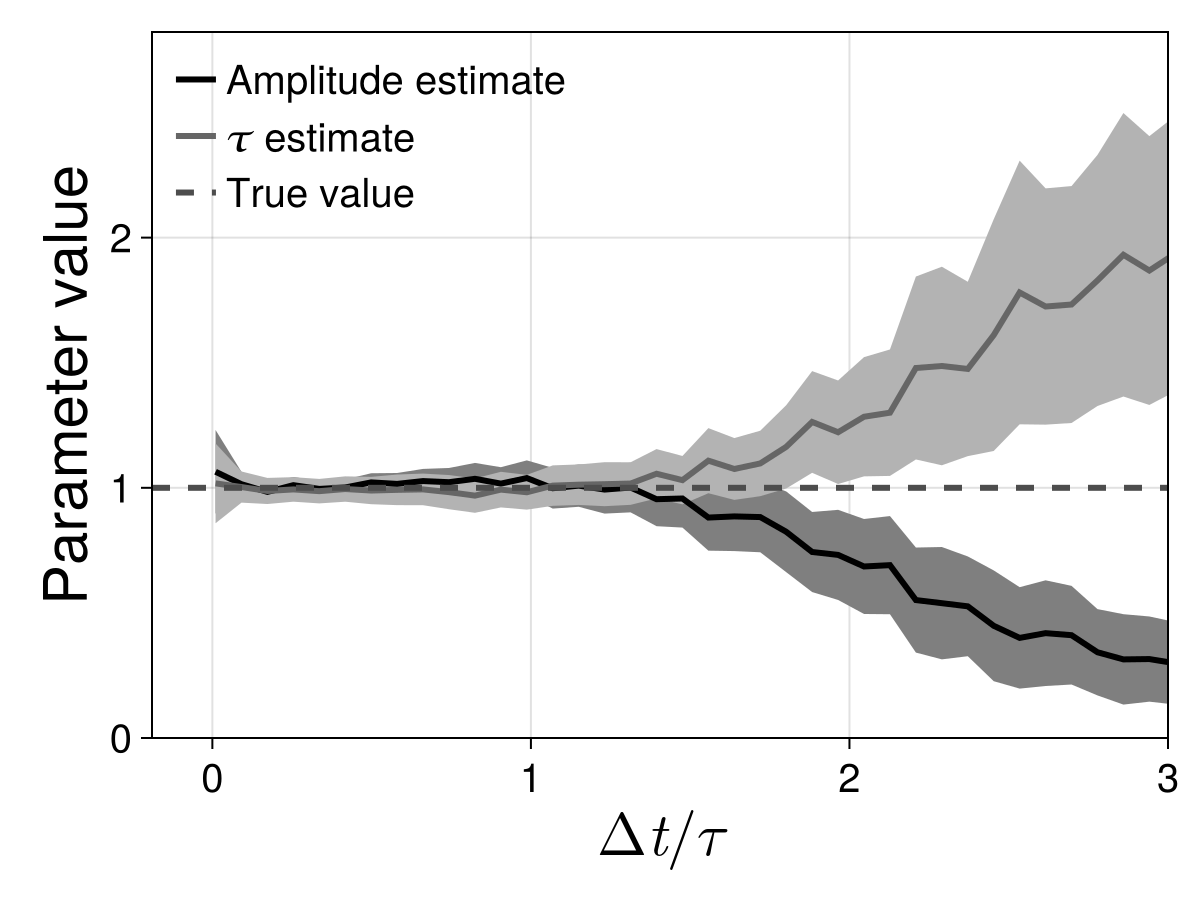}%
\quad
\includegraphics[height=4cm]{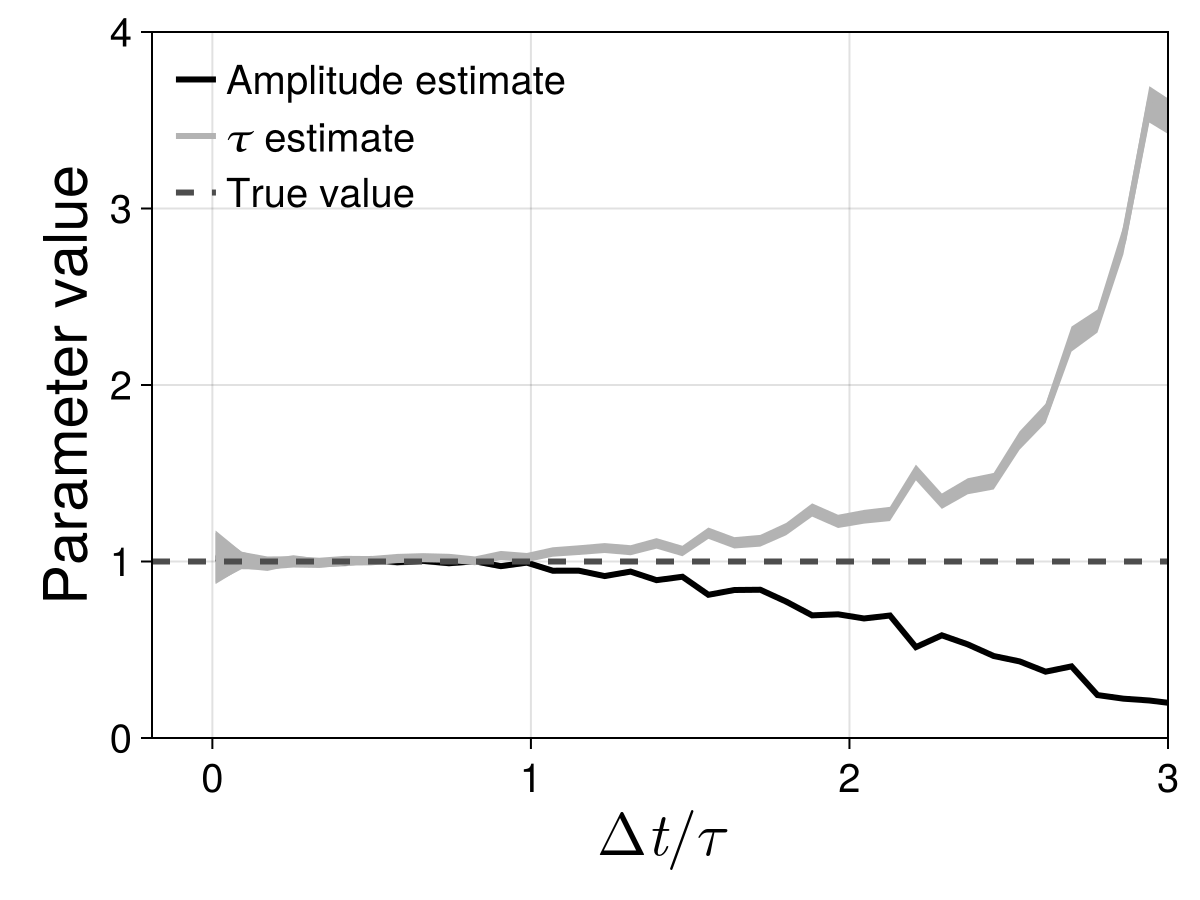}%
}
\caption{The parameter estimation as a function of the sample rate. The left graph was generated using Hamilton-Monte-Carlo with the NUTS sampler, and the right one using our EM algorithm. The ribbon around each is the standard deviation of the estimate, and for both parameters, the true values are equal to 1 (dashed line)}\label{fig:multitau}
\end{center}
\end{figure}

We validated our EM algorithm against Hamilton Monte-Carlo (HMC) methods using simulated ground truth OU time series with known parameters ($A=1$, and $\tau=1$). Our results can be applied to any parameter combination by rescaling the parameters. Hamilton MC methods draw samples directly from the posterior distribution, providing accurate estimations of the parameters and their uncertainties.  Specifically, we implemented the HMC methods in the Julia programming language \cite{bezanson2017julia} using the probabilistic programming package Turing.jl \cite{ge2018t}.  We obtained 2000 samples per data point employing the NUTS sampler \cite{RN24}.  The EM method outperforms the HMC method consistently by a factor of 10,000 using the same computer configuration as seen in Table \ref{table:benchmark}.  Fig. \ref{fig:multitau} illustrates that even though the EM method obtains the same parameter estimates, it severely underestimates their uncertainties.  This is not surprising since in EM, the uncertainty is estimated by assuming a posterior distribution with a Normal distribution.  In our earlier paper on parameter estimation of OU processes without measurement noise, we found that the maximum likelihood method similarly underestimated the uncertainty of the parameters by about a factor of five \cite{RN91}. In practice, we would recommend to run a few HMC analysis to establish the true uncertainties, and then compare them to the EM results.  In our experience, this method can establish a factor to describe how much the EM method underestimates the uncertainties and which can then be used in subsequent analysis.

\begin{table}
\centering
\caption{Benchmark Results: For the NUTS sampler we collected 2000 samples for thermal noise and 10000 samples for thermal and multiplicative noise to reach effective sample sizes larger than 400; The EM method was terminated when a relative tolerance of $10^{-6}$ was reached.  The computations were performed on a Macbook Pro M2 Max with 96GB of RAM.}
\begin{tblr}{
cell{1}{2} = {c=3}{},
vlines = {},
hlines = {},
}
& Execution Time & EM Method Execution \\
Chain length & NUTS Thermal & NUTS Thermal and Multiplicative & EM Method Execution  \\
100 & 64s & 402s & \\
500 & 485s & 2663s & 8.3ms \\
1000 & 804s & 5963s & 17.8ms \\
2000 & 2737s & 11142s & 26.5ms \\
5000 & & & 73.9ms\\
10000 & & & 136.2ms\\
\end{tblr}
\label{table:benchmark}
\end{table}

\begin{figure}[H]
\begin{center}
\includegraphics[width=0.5\textwidth]{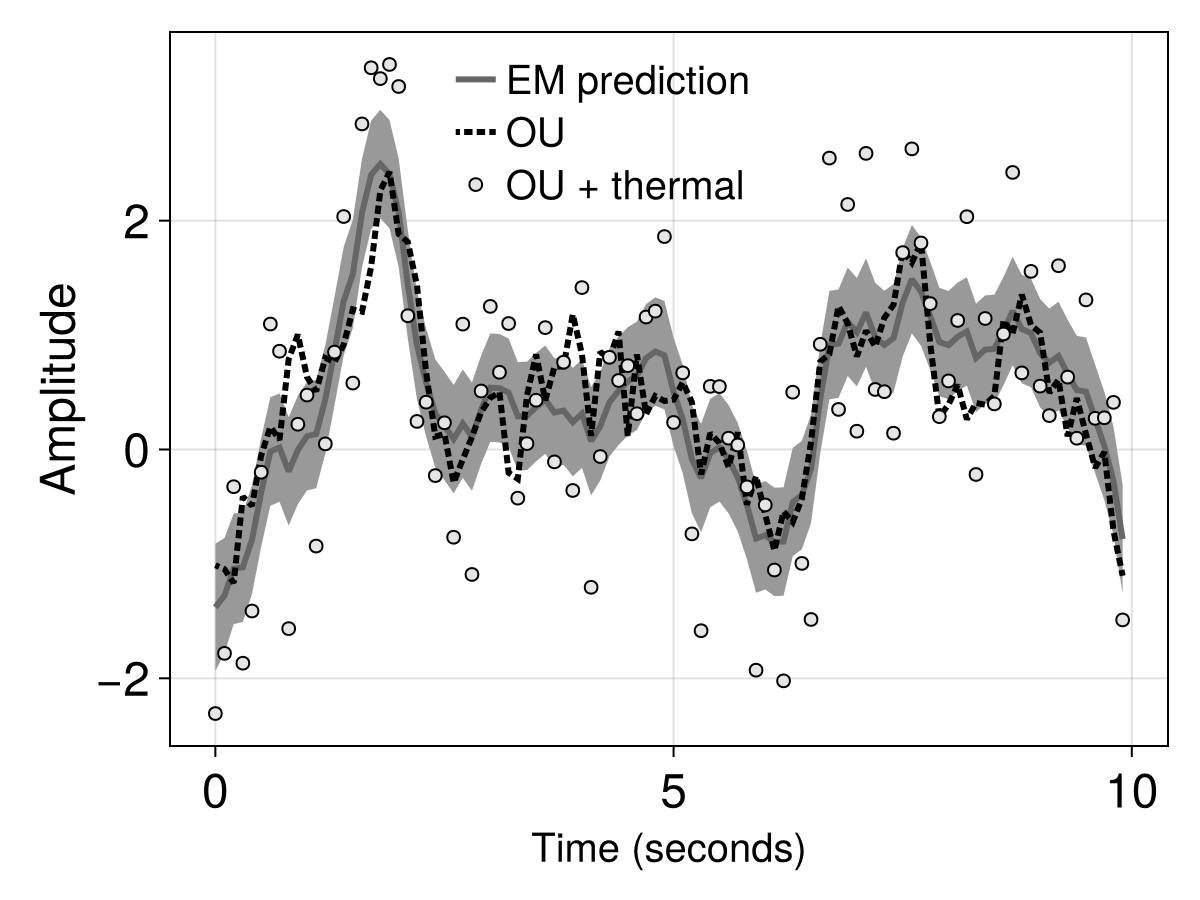}%
\caption{The Expectation Maximization (EM) fit for the first 100 points of an OU process obfuscated by thermal noise. The OU time series was created with parameters $A=1$ and $\tau =1$ at $\Delta t =0.1$ over a time interval of 100s. We then added thermal noise with variance 1. The Pearson correlation between the OU time series and the OU time series with added noise was 0.72.  The average of the posterior of the latent OU time series correlated with the original OU time series resulted in an improved Pearson correlation of 0.89.}\label{fig:emfit}
\end{center}
\end{figure}

Before we proceed with the mixed noise case, we would like to review alternative approaches to parameter fitting of Ornstein-Uhlenbeck processes.  The most commonly used method to extract parameters from an OU process is by fitting the power spectrum of the OU process to a Lorenzian, which is the Fourier transform of an exponential.  Power spectrum analysis is sometimes preferred in experimental situations where external noise sources at particular frequencies can be eliminated by adding additional fitting terms. Initially, these fits were performed using least-squares, but it soon became clear that the error distribution of the power spectrum is not Gaussian but rather exponential \cite{RN2}.  By taking this into account, power spectra can be fit using the maximum likelihood method that results in the correct estimation of parameters and their confidence intervals \cite{RN2, RN108}.  Once measurement noise is added to the signal, the fitting procedure based on likelihood becomes more complex since the Fourier contributions of the OU process draw from an exponential distribution and thermal noise draws from a $\chi^{2}$-distribution. The situation will become even more complex for multiplicative noise since, in Fourier space, the signal and the noise mix, which results in non-analytic distribution functions. As we argued before in \cite{RN91}, if done correctly, a maximum likelihood approach for time-series and power spectrum is mathematically equivalent and results in the same parameter and confidence interval estimation.  Where time-series analysis shines is the simultaneous estimation of the latent variables of the most likely $\{x_i\}$ (see Fig. \ref{fig:emfit}).  This information can be valuable for improving the estimation of the Pearson correlation between different OU processes with measurement noise that contributes to the variance but not to the cross-correlation and, therefore, reduces the correlation coefficient.  This issue is especially relevant in functional Magnetic Resonance Imaging (fMRI) data, where the correlation between brain regions is typically studied using Pearson correlations, and failure to remove noise results in an underestimation of the correlation coefficients\cite{RN44}.

\section{Properties of Multiplicative Noise}\label{mult}
In this section, we investigate the characteristics of multiplicative noise as defined by the parameter $\sigma_M$ in equation \ref{eqn:noise_label}. Compared to additive noise, multiplicative noise is more challenging to distinguish from the signal, and traditional methods such as Hamilton Monte Carlo (HMC) fail to sample the posterior distribution effectively, as we will show in the next section. This limitation can be explained by examining the power spectrum of the signal. Specifically, the first-order power spectrum shows that both thermal and multiplicative noise appear identical and uncorrelated, which could lead to the incorrect conclusion that both types of noise can be removed easily and treated similarly. However, upon examining higher-order spectra of the signals, a clear separation between thermal and multiplicative noise emerges, highlighting the difficulties of accurately separating and modeling multiplicative noise.
We can start to examine this relationship by defining the fourth-order correlation function
\begin{equation}
g_y^2(t_1,t_2) = <y^2_{t1} y^2_{t_2}> = E[y^2_{t1} y^2_{t_2}]= \int_{-\infty}^{\infty}\int_{-\infty}^{\infty}y^2_{t1} y^2_{t_2}\rho(y_{t_1}, y_{t_2}) \,dy_{t1}dy_{t2}
\end{equation}

Here $y_{t1}$ and $y_{t2}$ are the random variables whose probability distribution function (pdf) is equal to a Gaussian with variance equal to another variable $x_t$. We define another random variable $z$ such that, $z_{i}^2=x_{i}^2 - y_{i}$, where we use the fact that $E[x_i^2] = y_i$ for multiplicative noise. This gives us the expression
\begin{equation}
E[y^2_{t1} y^2_{t_2}] = [(z^2_{t1} + x_{t1})(z^2_{t2} + x_{t2})] = E[z^2_{t2}z^2_{t1} + z^2_{t2}x_{t1} + z^2_{t1}x_{t2} + x_{t2}x_{t1}]
\end{equation}

The middle two terms cancel using the law of total expectations since the expectation of $z_i^2$ is zero.
\begin{equation}
E[z^2_{t2}x_{t1}] = E[E[z^2_{t2}x_{t1} | x_{t1}x_{t2}]] = E[x_{t1}E[z^2_{t2}| x_{t1}x_{t2}]] = 0
\end{equation}

This leaves us with

\begin{equation}
g_y^2(t_1,t_2) = E[y^2_{t1} y^2_{t_2}] = E[z^2_{t2}z^2_{t1}] + E[x_{t2}x_{t1}]
\end{equation}

We see that the second-order correlation of the multiplicative noise is equal to that of the first-order correlation of the underlying process $x$ and that of a mean zero uncorrelated random variable $z$.

Assuming that $x_{t}$ is an Ornstein-Uhlenbeck process, we can show that the second-order correlation is the same shape as the first-order correlation and, as a result, show that the second-order spectrum of multiplicative noise is hidden by this process. The math is included in the appendix, with the final result being

\begin{equation}\label{eq:sum_exp}
g_x^2(t_1,t_2) = \frac{\sigma^{4}e^{-2\theta(s+t)}}{2\theta}[\frac{3}{2\theta}e^{4\theta min(s,t)} -2e^{2\theta min(s,t)} + 1 \\ + (e^{2\theta min(s,t)}-1)(e^{2\theta max(s,t)} - e^{2\theta min(s,t)})] + ...
\end{equation}

The final result is a sum of exponentials. Thus, the power spectrum will be the sum of Lorentzian due to the linear properties of the  Fourier transform, which have the same general shape as the multiplicative noise previously shown.

\begin{figure}[htbp]
    \centering
    \begin{minipage}[b]{.3\linewidth}
        \centering
        \textsf{\textbf{Time series}}
    \end{minipage}%
    \begin{minipage}[b]{.3\linewidth}
        \centering
        \textsf{\textbf{Power Spectrum}}
    \end{minipage}%
    \begin{minipage}[b]{.3\linewidth}
        \centering
        \textsf{\textbf{Power Spectrum (second order)}}
    \end{minipage}
    
    \par\bigskip
    \begin{minipage}[b]{.3\linewidth}
        \centering
        \includegraphics[width=\linewidth]{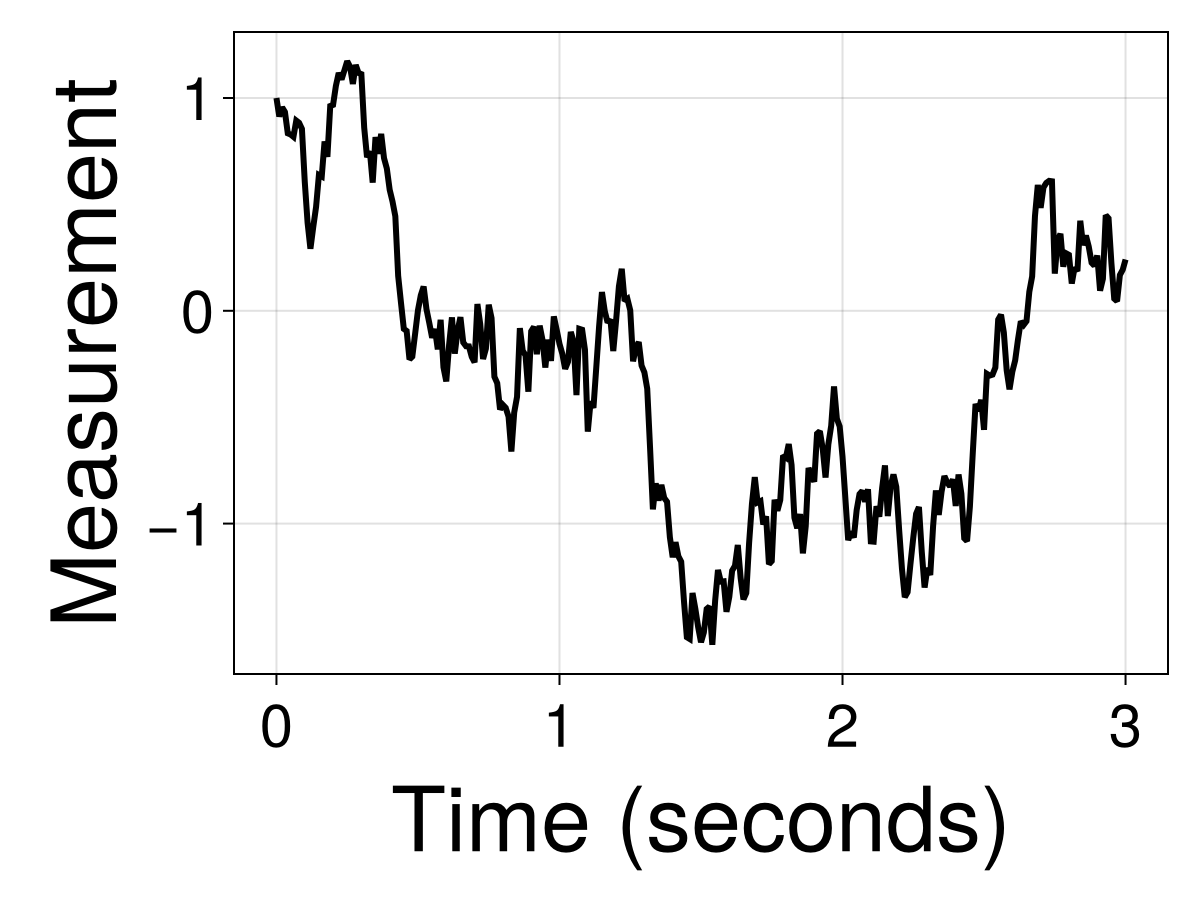} 
    \end{minipage}%
    \begin{minipage}[b]{.3\linewidth}
        \centering
        \includegraphics[width=\linewidth]{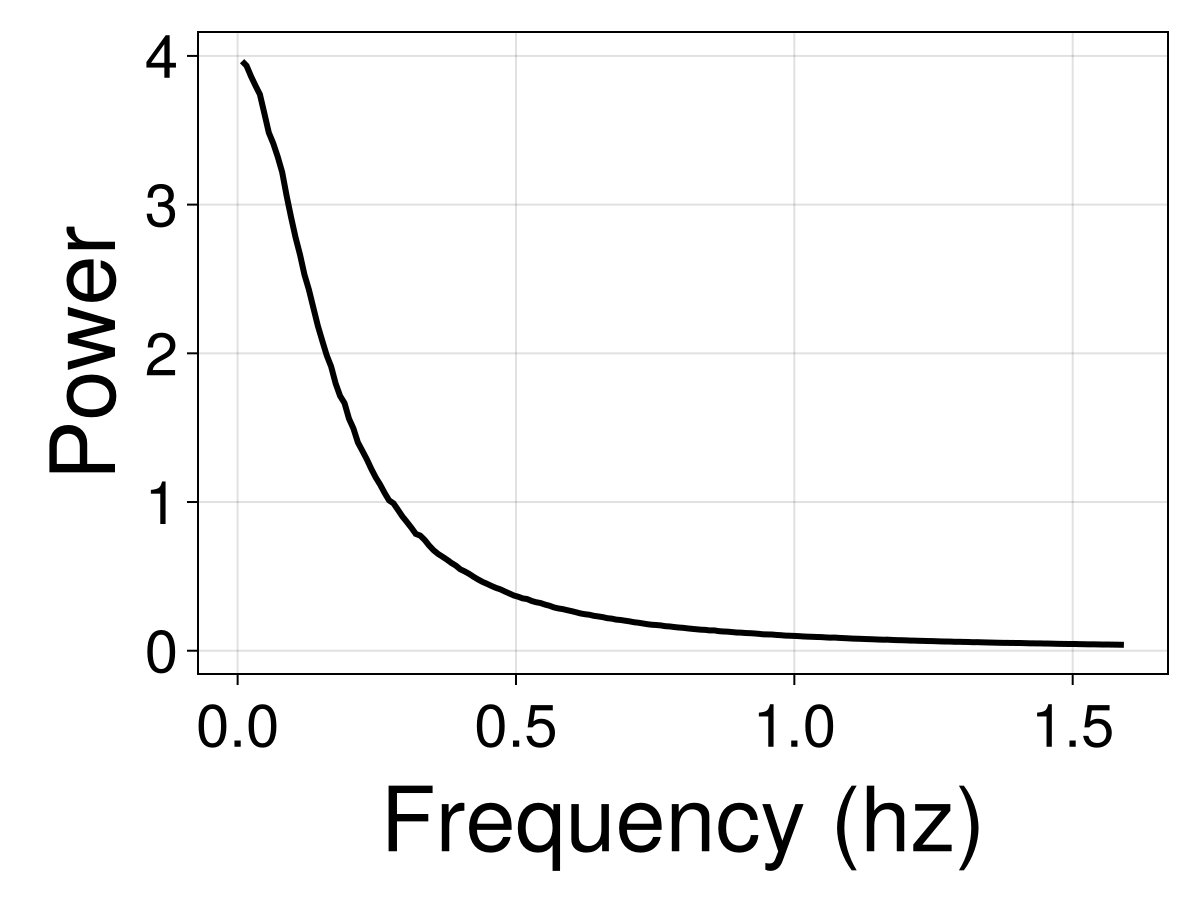} 
    \end{minipage}%
    \begin{minipage}[b]{.3\linewidth}
        \centering
        \includegraphics[width=\linewidth]{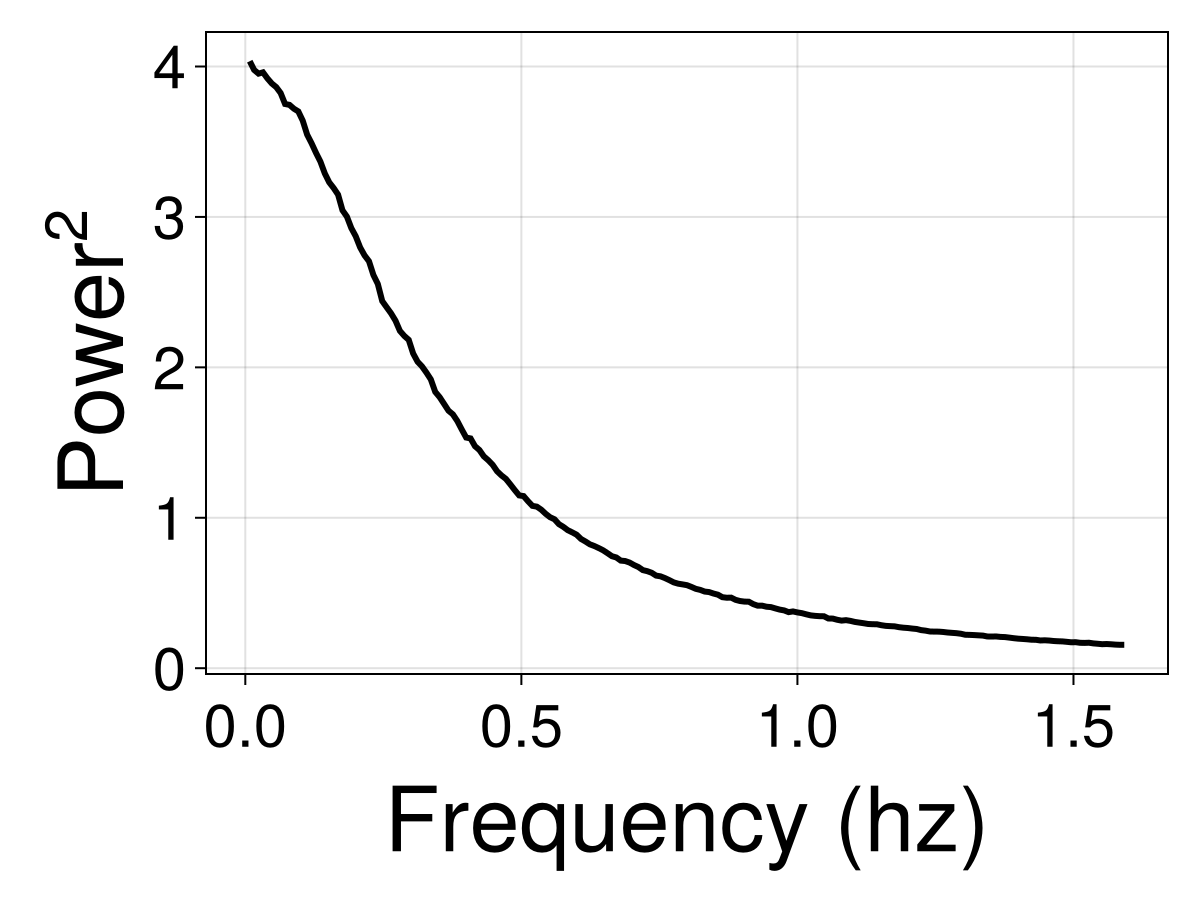} 
    \end{minipage}
    
    \par\bigskip
    \begin{minipage}[b]{.3\linewidth}
        \centering
        \includegraphics[width=\linewidth]{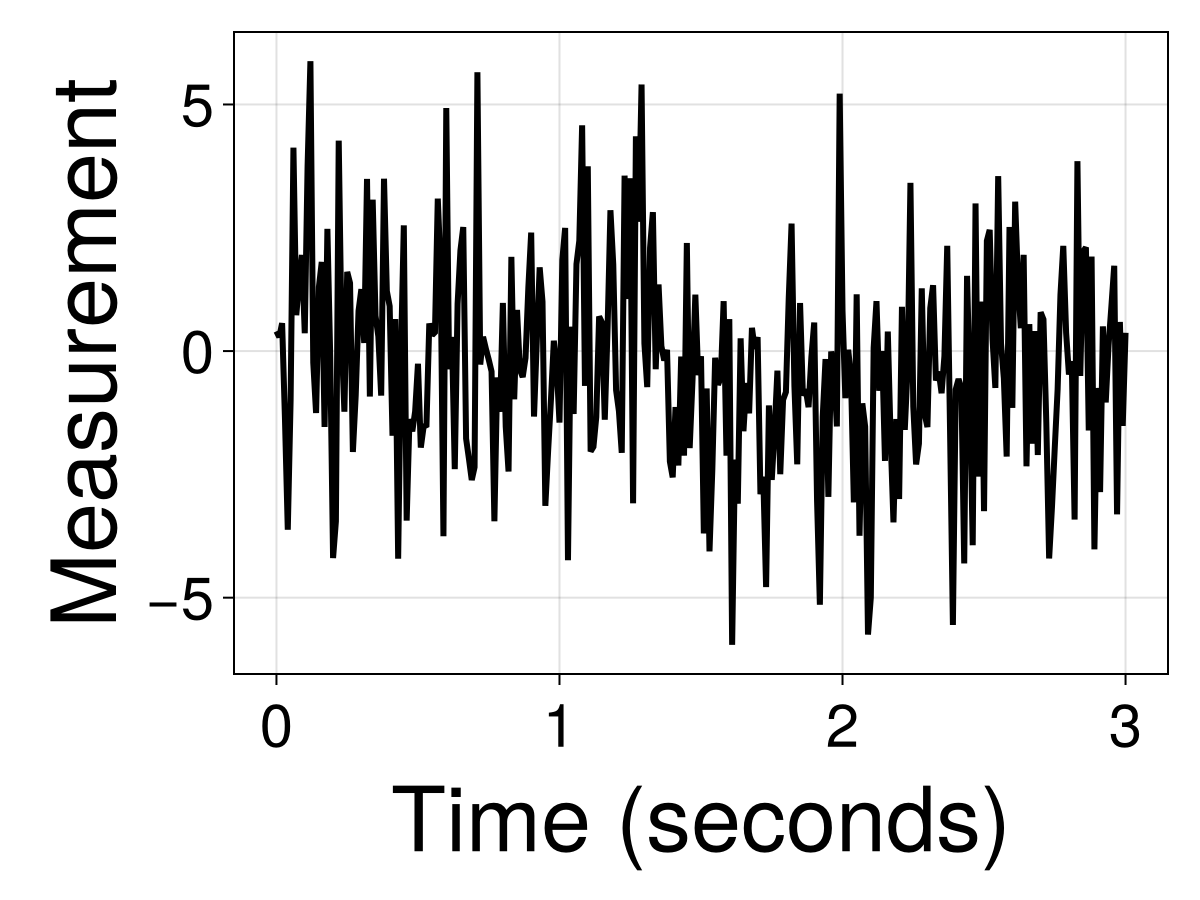} 
    \end{minipage}%
    \begin{minipage}[b]{.3\linewidth}
        \centering
        \includegraphics[width=\linewidth]{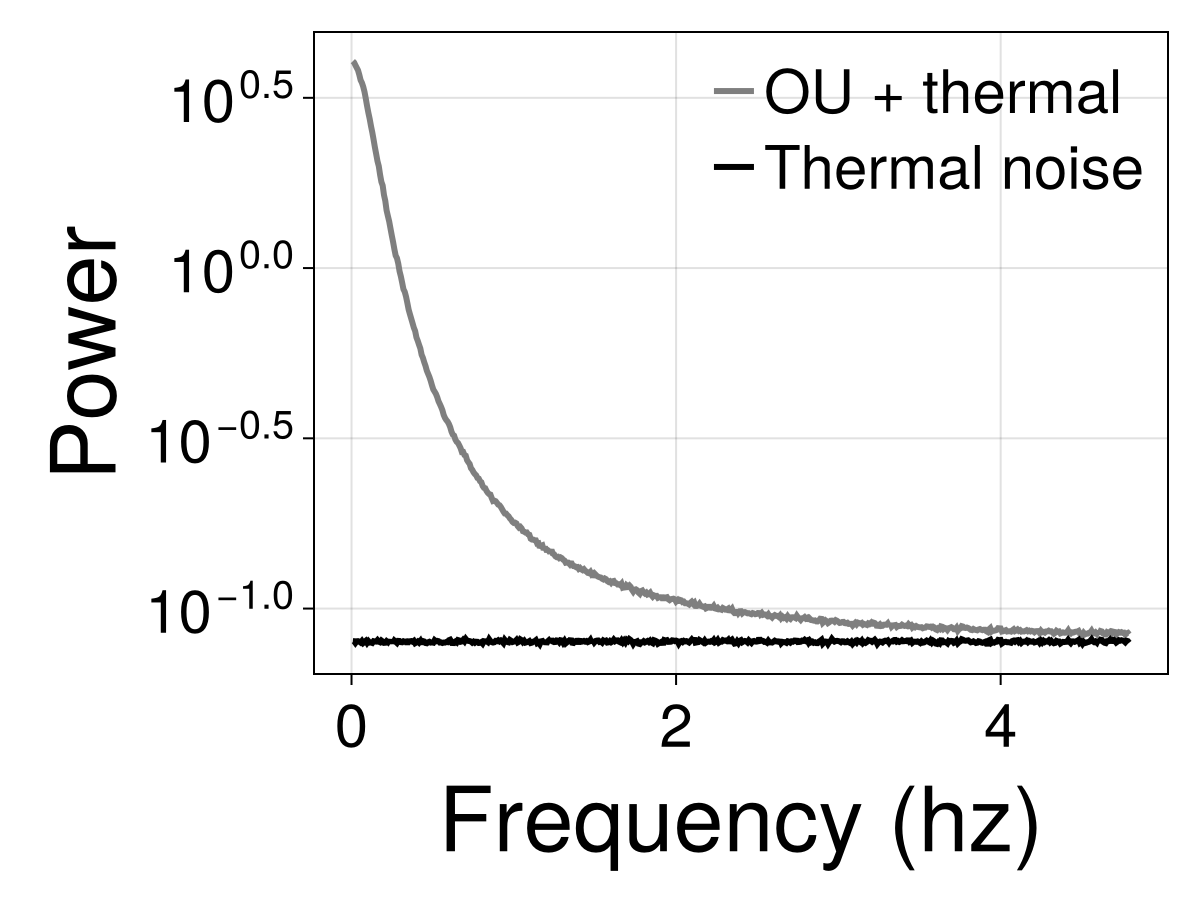} 
    \end{minipage}%
    \begin{minipage}[b]{.3\linewidth}
        \centering
        \includegraphics[width=\linewidth]{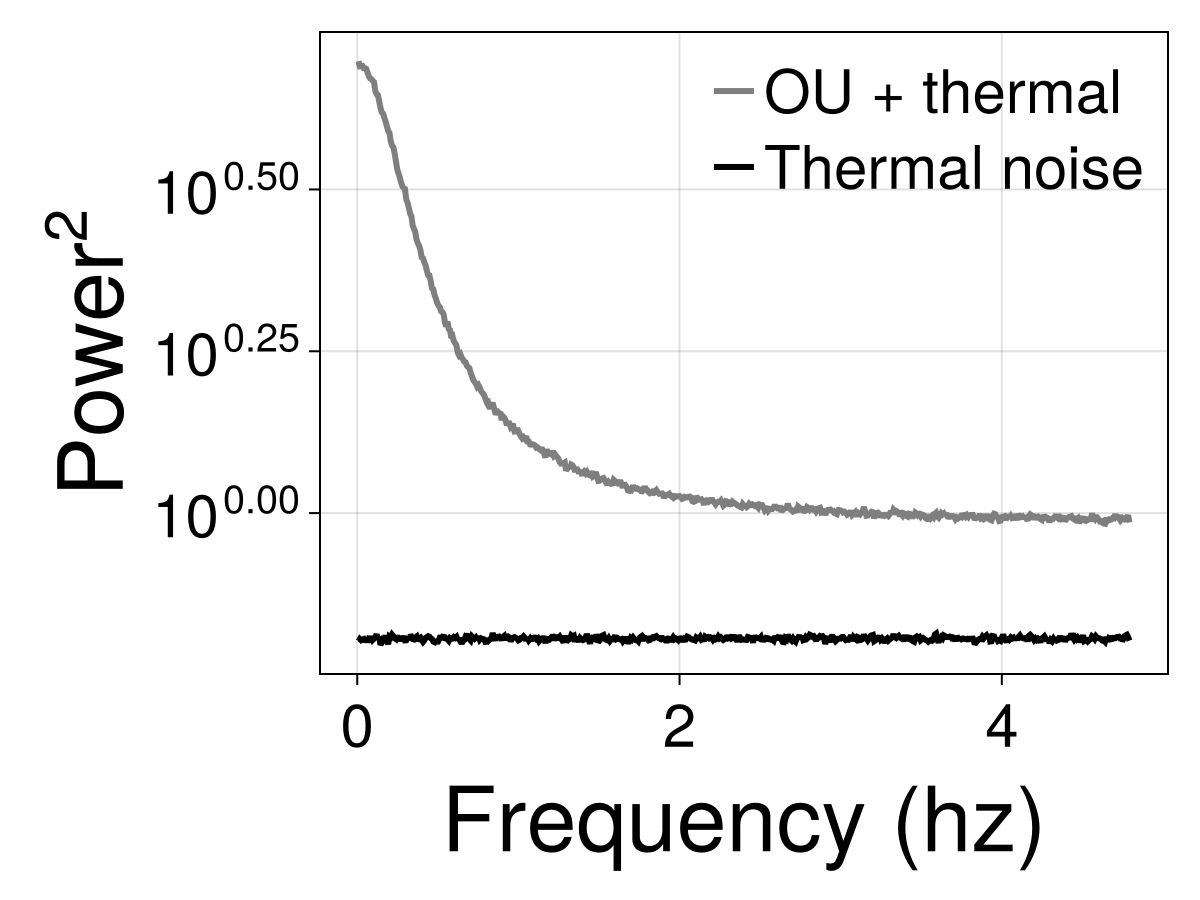} 
    \end{minipage}
    
    \par\bigskip
    \begin{minipage}[b]{.3\linewidth}
        \centering
        \includegraphics[width=\linewidth]{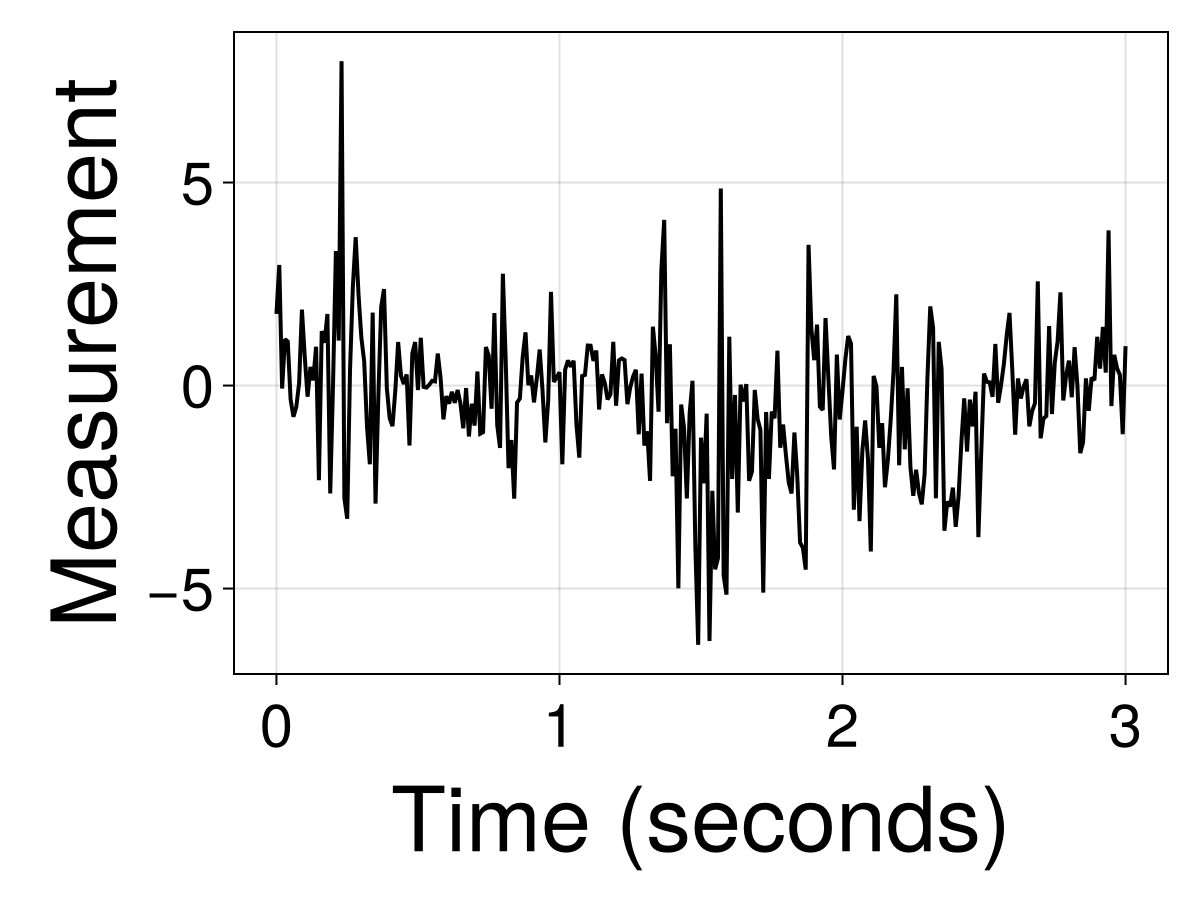} 
    \end{minipage}%
    \begin{minipage}[b]{.3\linewidth}
        \centering
        \includegraphics[width=\linewidth]{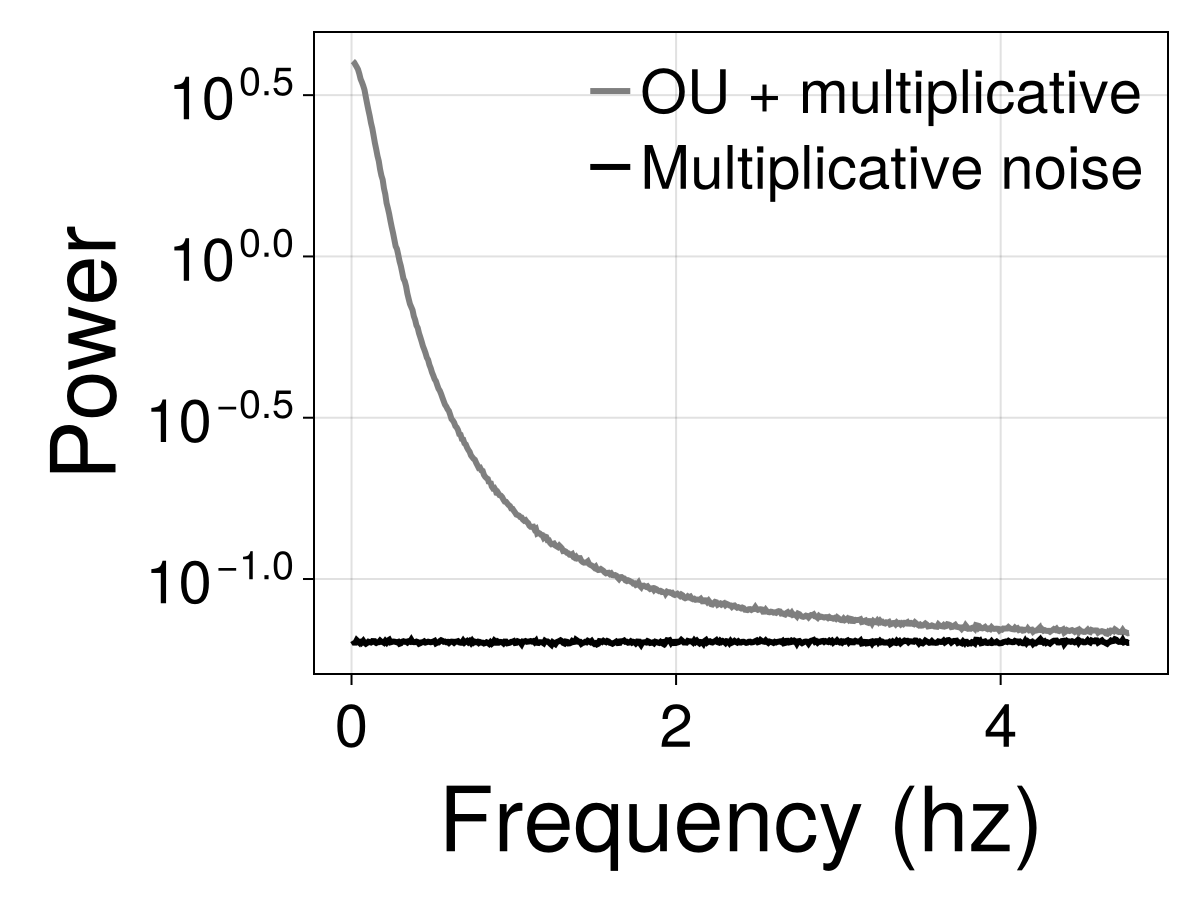} 
    \end{minipage}%
    \begin{minipage}[b]{.3\linewidth}
        \centering
        \includegraphics[width=\linewidth]{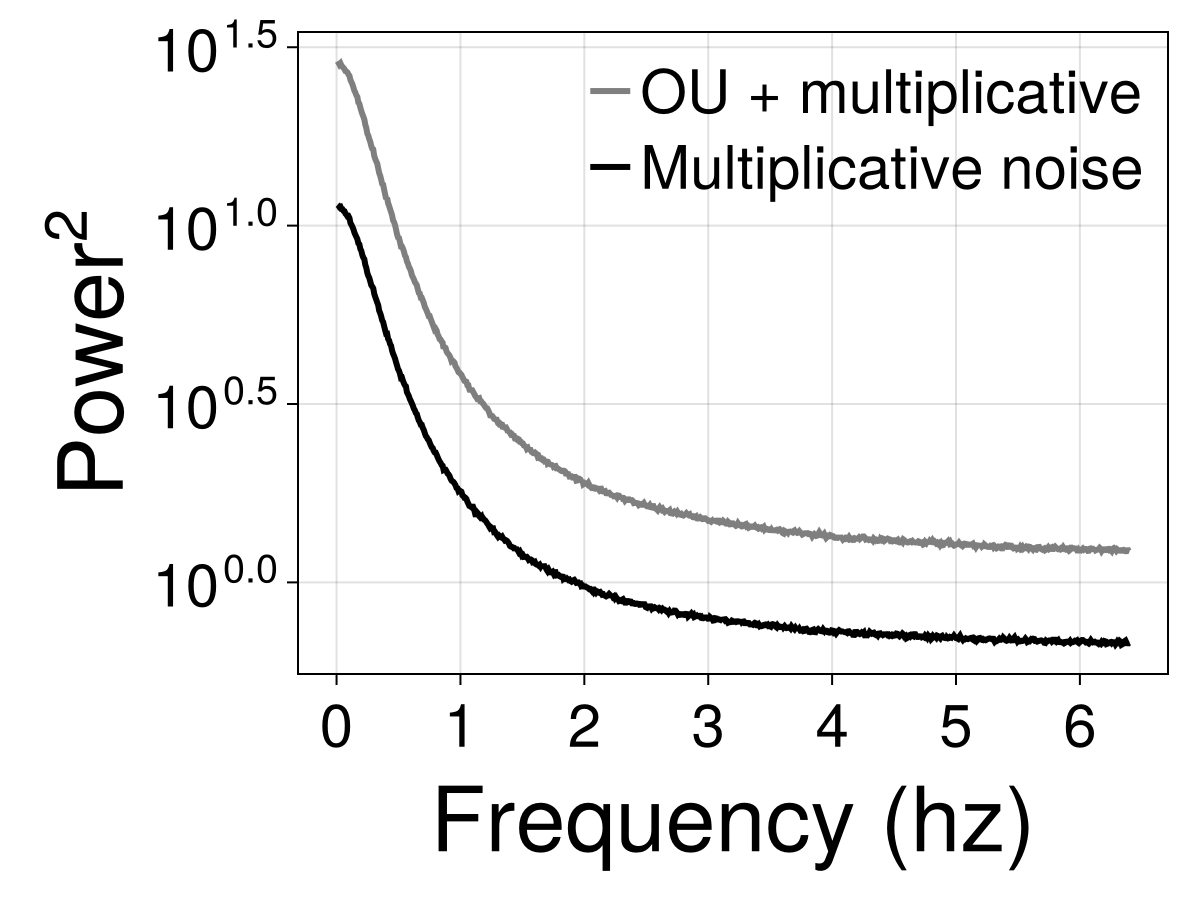} 
    \end{minipage}
    
    \caption{Power spectra of the Ornstein-Uhlenbeck process with different types of noise added.
Fig. A is the baseline Ornstein-Uhlenbeck signal. Thermal noise (Fig. B) 
adds a constant in the power spectrum.  Multiplicative noise (Fig. C) closely resembles the
Ornstein-Uhlenbeck signal in the second-order power spectrum, while it looks thermal in the first order. For the
power spectrum, the curves have been normalized such that the area under the curve is equal to the variance of the
signal. For the second-order power spectrum, this is equivalent to the variance of the signal squared.}
    \label{fig:three graphs}
\end{figure}

Fig. \ref{fig:three graphs} shows simulated OU time series with added thermal and multiplicative noise with their respective first and second-order power spectra, confirming our theoretical calculations. 

\section{Parameter Fitting with added Multiplicative Noise}\label{Mult_val}
In this section, we employ the Julia programming language \cite{bezanson2017julia} to investigate the effectiveness of Hamilton Monte Carlo (HMC) and probabilistic programming in distinguishing multiplicative and thermal noise from an Ornstein-Uhlenbeck (OU) signal. To test our fitting algorithm, we generate OU time series contaminated with varying levels of thermal and multiplicative noise, and sample from the posterior distribution using the No U-Turn Sampler (NUTS) \cite{RN24} implemented in the probabilistic programming package Turing.jl \cite{ge2018t}. We assume uniform prior distributions for all parameters. More details on the algorithm and additional plots and data can be found on our \href{https://github.com/lcneuro/Ornstein_Uhlenbeck}{GitHub repository}.

\begin{figure}[h]
\begin{center}
\includegraphics[width=0.4\textwidth]{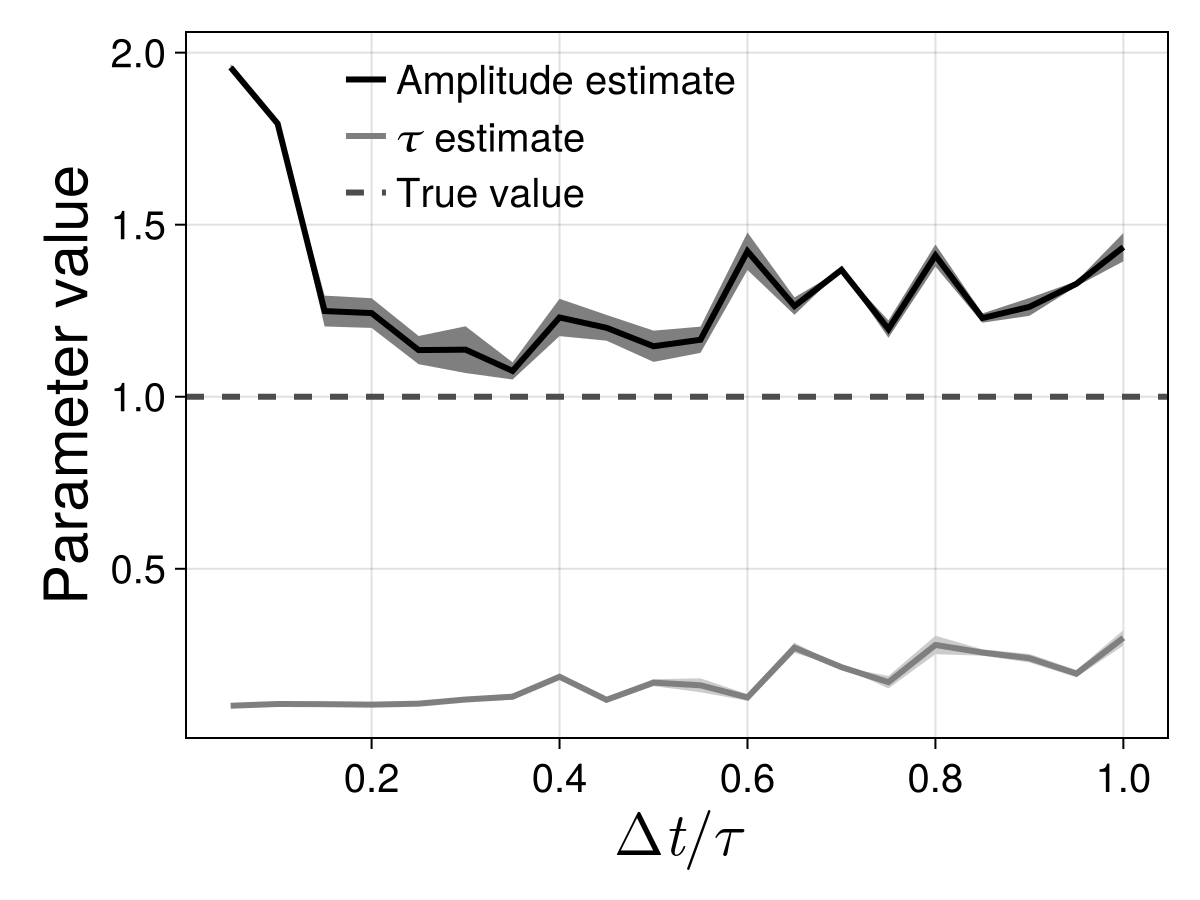}
\caption{Parameter estimation for pure multiplicative noise using HMC. We cannot estimate the true values for the amplitude and tau of the OU signal for any sample rate tested.}\label{fig:mult_fail}
\end{center}
\end{figure}

Our results show that our fitting algorithm struggles to distinguish between the OU process and multiplicative noise for any sampling time. This is likely due to the similar signature of both processes. In contrast, thermal noise is shown to be easily distinguishable from the underlying signal when sampling below the time constant $\tau$. We demonstrate this in Fig. \ref{fig:mult_fail} by applying our algorithm for a range of sampling times, each consisting of 1500 points. The generated OU signals all have zero drift and amplitude, and $\tau$ is equal to one.

\begin{figure}[h]
\begin{center}
\includegraphics[width=0.5\textwidth]{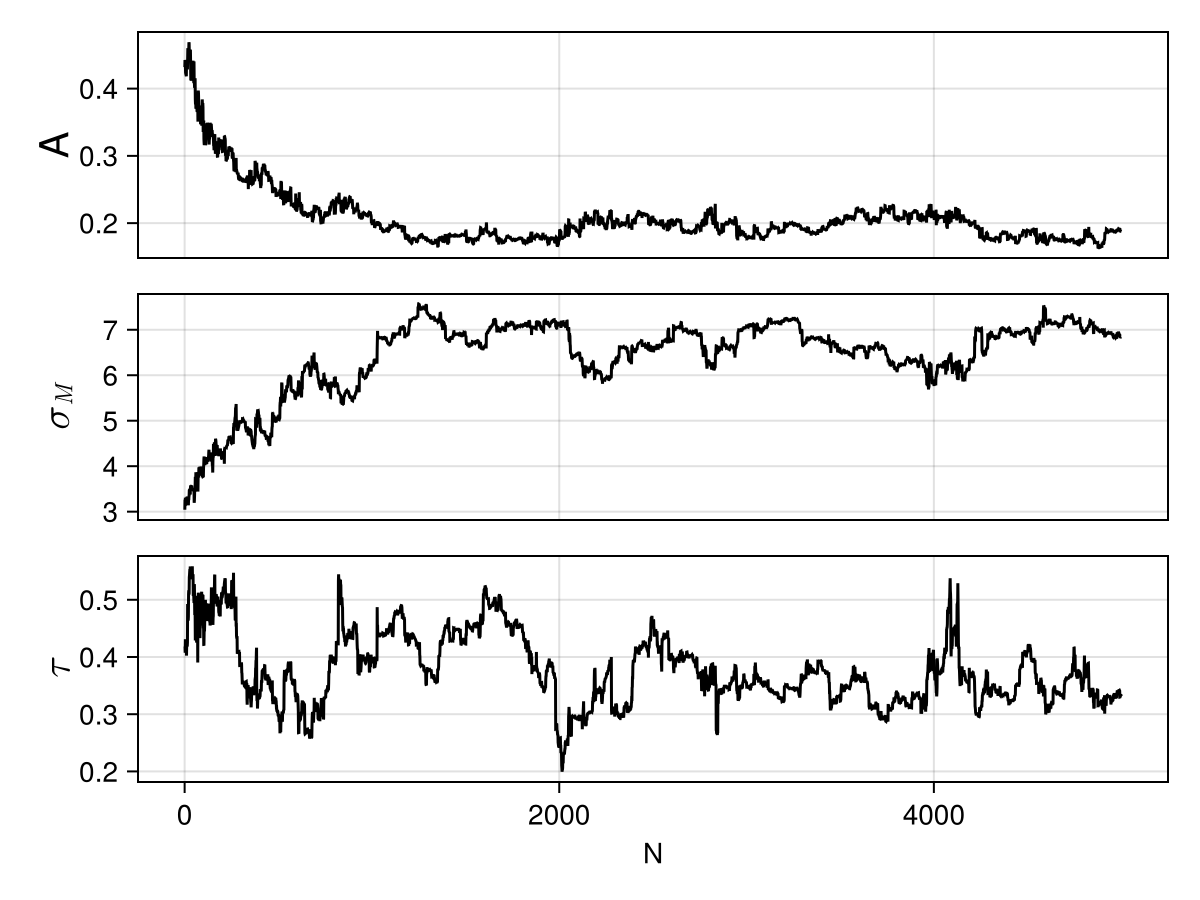}
\caption{A typical single NUTS chain on an OU process with added multiplicative noise. We simulated an OU process with $A=1$ (var OU) and $\tau =1$ over 250sec at $\Delta t = 0.5$. We then added multiplicative noise with $\sigma_{M} = 0.2$. 
 The figure shows the MCMC chain for A, $\sigma_M$ and $\tau$.  }\label{fig:nuts_chain}
\end{center}
\end{figure}

Analysis of the HMC chains shows additional difficulties with using traditional Monte Carlo methods for fitting the posterior with multiplicative noise. Fig. \ref{fig:nuts_chain} shows the samples from a Monte Carlo chain. We see that the chains do not converge to their true values, but consistently drift instead to a different local minimum with unrealistic parameter values.  Similar posterior probability pathologies have been observed in hierarchical models \cite{betancourt2013hamiltonianmontecarlohierarchical}, which were first described in \cite{10.1214/aos/1056562461} using a funnel distribution. Surprisingly, the variance that is predicted by the parameters is an order of magnitude of the true variance.  For example in Fig. \ref{fig:nuts_chain} we observe a relatively stable region with $A\approx 0.2$ and $\sigma_{M} \approx 7$.  The total variance of OU process plus noise is calculated as $ A(1 + \sigma_{M}^2) \approx 10$, whereas the true variance of the simulated time series is $1.2$.  Additionally, when running the multiplicative noise posterior model with OU data containing no multiplicative noise, the HMC sampler reproducibly drifts into spaces, indicating that the signal is dominated by multiplicative noise.

Despite the previous results, it is possible to distinguish multiplicative noise from the signal under certain conditions. Most measurements will have a mixture of both types of noise, and we found that if the ratio of multiplicative to thermal noise is known, then it is possible to estimate both the thermal and multiplicative noise.  In the following, we assume we know the thermal to multiplicative noise ratio. We define the ratio of thermal to multiplicative noise as,
$E[x^2_i] \sigma^2_M/ \sigma^2_N$. Since the variance of the multiplicative noise is not constant and is signal-dependent, we use the expectation of the signal's variance in the following.

Here we imagine experimental scenarios where one can determine thermal noise and multiplicative noise independently and, from that, establish their ratio.  Often, multiplicative noise gets introduced by counting processes (Poisson distributions whose variance is equal to the mean), whereas thermal noise could be introduced by noisy amplifiers.  In functional magnetic resonance neuroimaging, for example, it is possible to determine this ratio using a dynamic fMRI phantom \cite{KUMAR2021117584}.

\begin{figure}[h]
\centering
\begin{subfigure}{.49\linewidth}
    \centering
    \includegraphics[width=.7\linewidth]{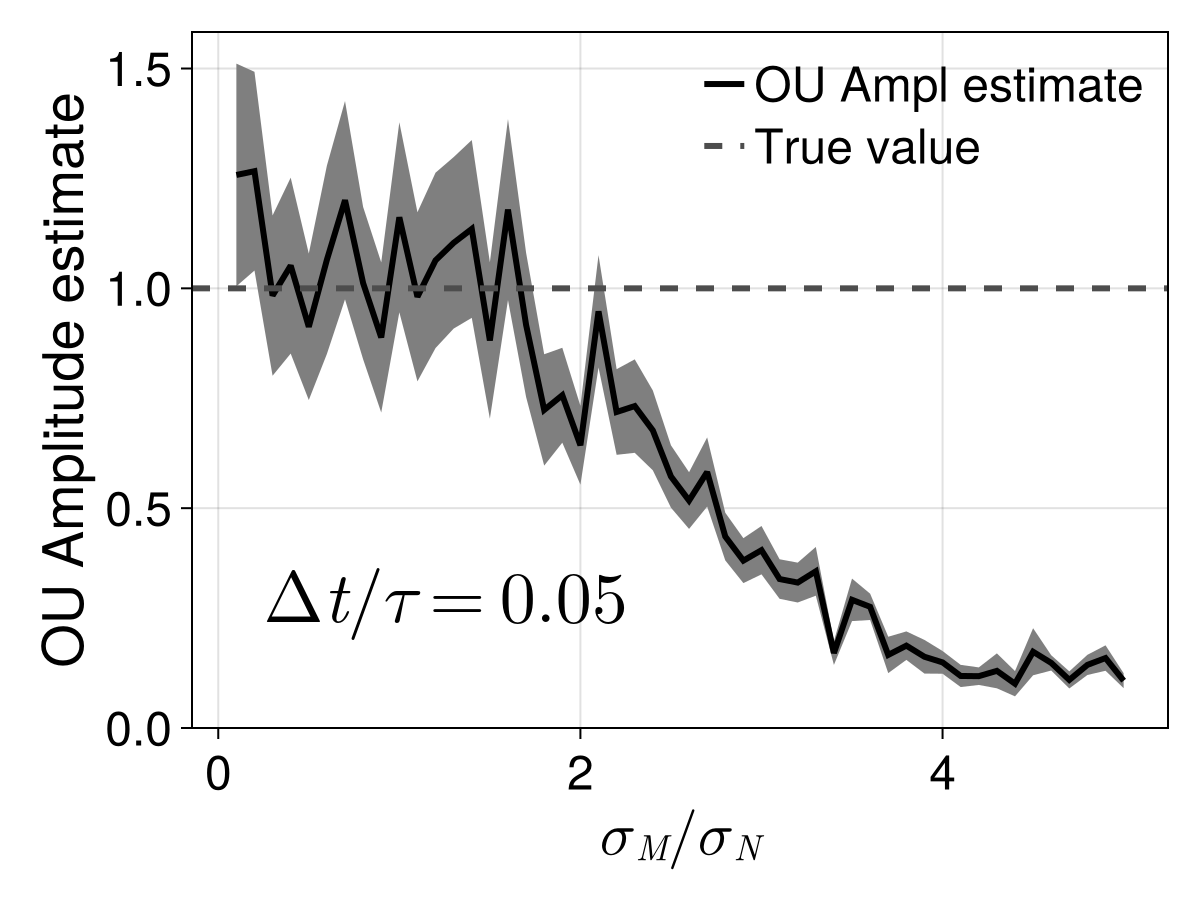}  
    \label{SUBFIGURE LABEL 1}
\end{subfigure}
\begin{subfigure}{.49\linewidth}
    \centering
    \includegraphics[width=.7\linewidth]{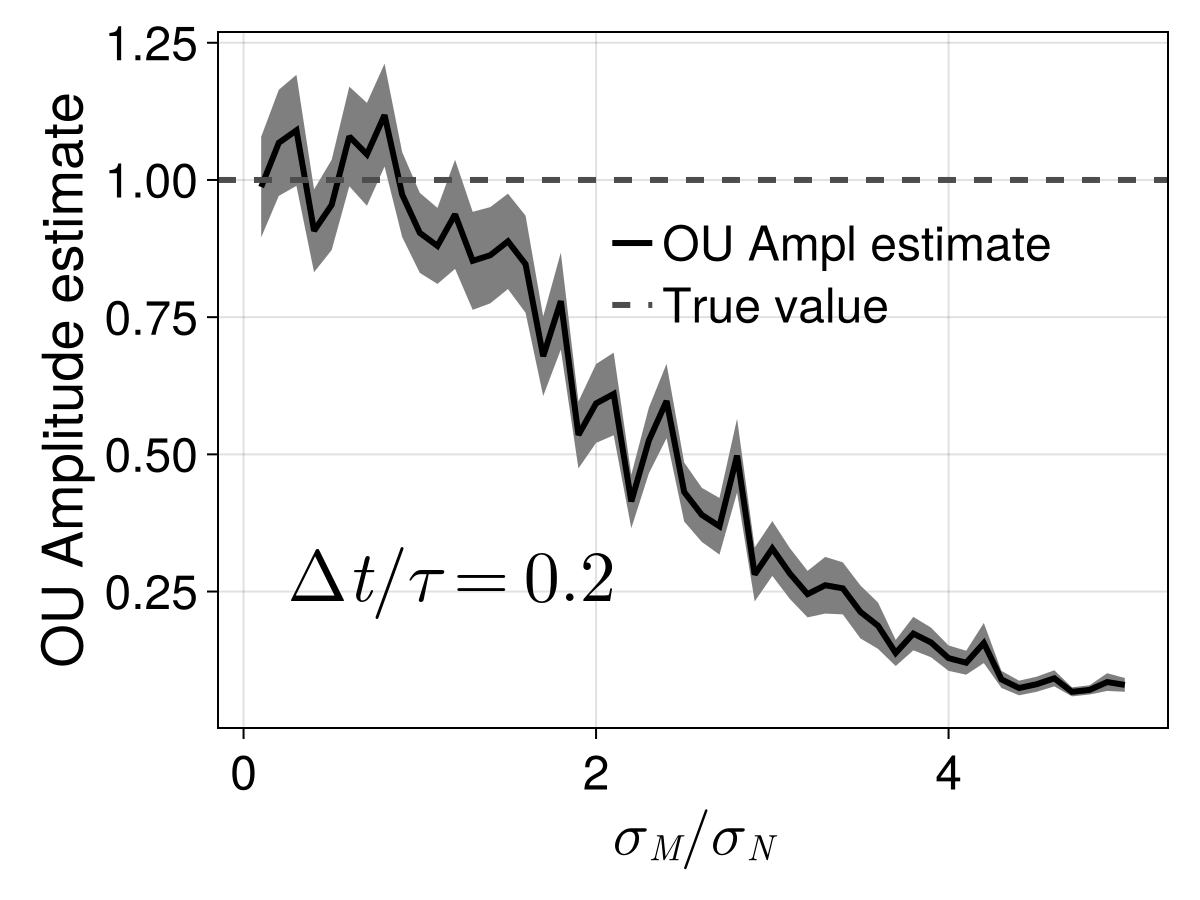}
    \label{SUBFIGURE LABEL 2}
\end{subfigure}
\begin{subfigure}{.49\linewidth}
    \centering
     \vspace{1cm}
    \includegraphics[width=.7\linewidth]{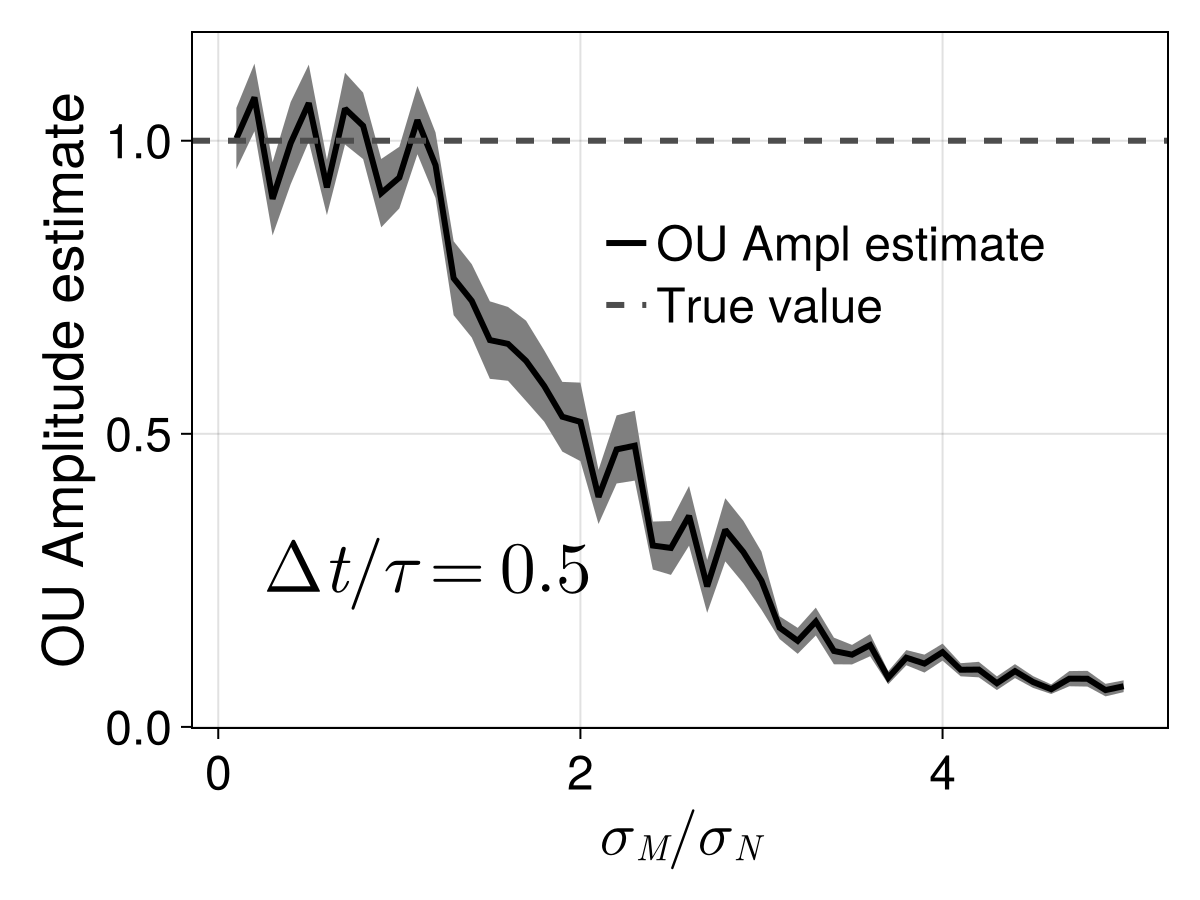}  
    \label{SUBFIGURE LABEL 3}
\end{subfigure}
\begin{subfigure}{.49\linewidth}
    \centering
     \vspace{1cm}
    \includegraphics[width=.7\linewidth]{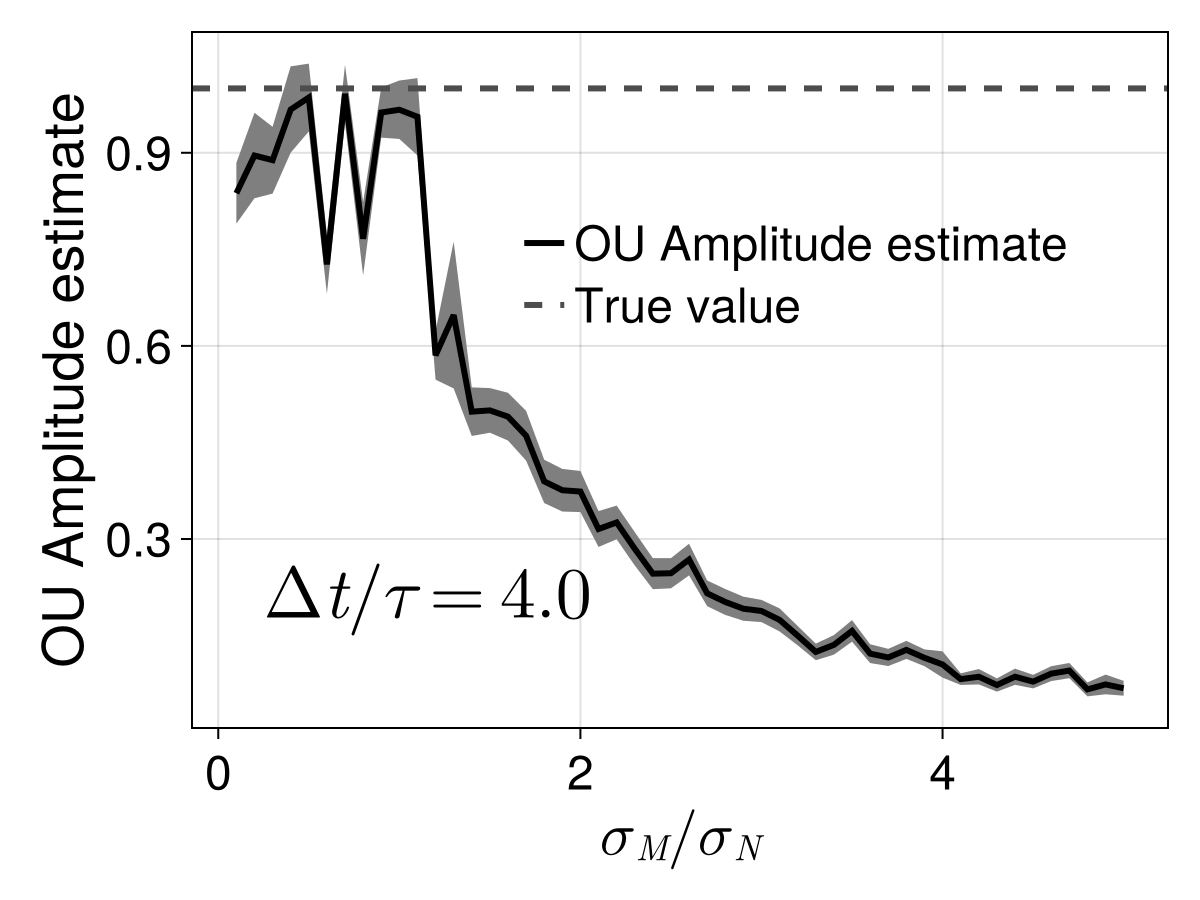}  
    \label{SUBFIGURE LABEL 4}
\end{subfigure}
\caption{OU Amplitude estimate for different thermal to multiplicative noise ratios employing a NUTS sampler. Each figure represents a different $\Delta t$ from 0.05$\tau$ to 4$\tau$. The true value for $A$ is represented by a dashed line. The total noise variance is normalized to a magnitude of 0.2 for each ratio. For each data point, we simulated 1500 time points and drew 1000 samples from the NUTS sampler at 0.65 acceptance rate.  As the ratio of multiplicative to thermal noise increases, we see the amplitude is consistently underestimated, with this effect becoming more pronounced at higher sampling rates.  This underestimation of $A$ is accompanied by a decrease in the effective sampling size (ESS) (see Discussion). At $\sigma_{M}/\sigma_{N} < 1.5$, we observe an ESS over 500, with a strong decrease in ESS beyond 1.5.  This decrease in ESS is an indication of an ill-formed posterior.}
\label{amp_est}
\end{figure}

We now investigate under what circumstances the fitting algorithm can resolve the signal from thermal and multiplicative noise. From our previous findings, we expect that there will be a transition point since pure multiplicative noise cannot be separated, whereas thermal noise can be easily separated. Indeed, when we vary the multiplicative to thermal noise ratio, we observe a plateau and transition ratio as illustrated in Fig. \ref{amp_est}.

We find that this plateau and transition ratio depends on the data sampling rate in relationship to the OU relaxation rate $\tau$.  This is not surprising since the only way to distinguish an OU process from thermal noise is to sample faster than the OU relaxation rate (see Fig.~\ref{fig:multitau}).  Once we can distinguish the OU process from thermal noise, we can determine the multiplicative noise from the known ratio of multiplicative to thermal noise, leading to a successful separation of noise from OU signal.  In Fig. \ref{amp_est}, we plot the measured OU amplitude (true amplitude is 1) versus the multiplicative to thermal noise ratio.  For low ratios, we observe a plateau over which the separation of noise and signal is successful.  Once we exceed that ratio, the separation fails.  When varying the sampling rate given as the ratio of $\Delta t/\tau$, we observe that the plateau shrinks as we decrease the sampling rate. It is worth noting that, during all the simulations, fitting with multiplicative noise took substantially longer than with thermal noise when using HMC sampling methods (see the Discussion section for more detail).

\section{Discussion}
In the context of an Ornstein-Uhlenbeck (OU) process, we have developed an efficient algorithm for fitting an OU signal with thermal noise and investigated the properties of multiplicative noise. In cases where the multiplicative signal is insignificant or nonexistent, the expectation maximization (EM) algorithm can be more practical in large-scale simulations. The EM algorithm does not necessarily improve the overall accuracy of the fit as compared to HMC. Still, it can facilitate the analysis of larger datasets, which are increasingly common in biological applications, and pose a struggle for MCMC-based simulations \cite{Bardenet, sejnowski_putting_2014}.  However, the assumption of an underlying OU function with additional uncorrelated noise limits the applicability of our algorithm to a particular class of fitting functions. In addition, we can see from Fig. \ref{fig:multitau} that the EM algorithm underestimates the uncertainty of our fit, which can lead to unfounded confidence in a particular parameter fit. HMC, on the other hand, is slower but supports a broad range of fitting functions, including those that incorporate multiplicative noise while maintaining more reasonable levels of uncertainty for our parameter estimation. The question then arises regarding how to fit large datasets with multiplicative noise. Our algorithm's effectiveness is due to the OU process's Gaussian nature and underlying thermal noise, which allows the likelihood to factor. The model's validity could be further confirmed by comparing it with other approximate methods, such as Automated Differentiation Variational Inference (ADVI) \cite{https://doi.org/10.48550/arxiv.1603.00788}. In future research, it would be interesting to explore whether our algorithm can be extended to handle multiplicative noise and be applied to other types of signals, as the ability to handle various kinds of noise will be critical in making accurate predictions in real-world applications.

To interpret Monte-Carlo methods proper convergence is essential. One standard metric used in the field is the effective sampling size (ESS).  The ESS measures how well a Monte Carlo chain is sampling the posterior distribution by computing the effective sample size of independent samples \cite{Vehtari_2021}. Although there are no rigid guidelines as to what value of ESS is acceptable, an ESS greater than 400 is generally preferred. During our analysis, we noticed slow ESS convergence, which was the primary reason for the long sampling time necessary to reach this threshold. This tends to colloquially be due to either multimodal posterior distributions or when a chain is stuck in a region of high curvature \cite{Vehtari_2021}. We consistently observed very slow ESS rates for multiplicative noise, and analyzing the correlational values for amplitude and the coefficient for multiplicative noise, we see a very high negative correlation, suggesting potential degeneracy in the parameters. This is further supported by the shape of chain convergence shown in Fig. \ref{fig:nuts_chain}. HMC converges to a set of parameters that provide similar variance to the underlying signal but with the wrong parameters. This suggests that adding multiplicative noise modifies the distribution such that the posterior no longer converges to the correct parameters.  Further research is necessary to identify this pathology in detail and to potentially develop a specialized HMC sampler that can properly sample from these distributions as described in \cite{betancourt2013hamiltonianmontecarlohierarchical}.

In the case of pure multiplicative noise, we have demonstrated that HMC struggles to differentiate between multiplicative noise and the underlying signal. Intuitively, one would think that since the magnitude of the multiplicative noise follows the signal, there should be some distinguishing signatures visible in the power spectrum, but as our theoretical analysis shows, the difference between thermal and multiplicative noise is only visible in the second-order power spectrum. This suggests that first-order statistics are inadequate to describe the system fully and that an Ornstein-Uhlenbeck process with multiplicative noise is no longer Gaussian. This issue is especially relevant in fMRI data, where the correlation between brain region neural activation time series is studied, and failure to remove noise results in an underestimation of correlation \cite{RN44}.

The selection of an appropriate sampling rate is a critical factor in experimental settings to minimize noise and effectively distinguish the desired signal. For pure thermal noise, we observe that as the sampling rate increases, we lose the ability to differentiate between thermal noise and the underlying signal. Given a specific time constant, experimenters can use our results to optimize the sampling rate for better data processing. However, this optimal sampling rate becomes more complex with added multiplicative noise. It depends on the thermal to multiplicative noise ratio within the time series and the number of samples. We discovered that as the ratio of multiplicative noise increases relative to thermal noise, there is a tighter requirement for a higher sampling rate and, consequently, more samples. This may pose a challenge in specific experiments with a limitation on the sampling rate. Nevertheless, we have identified the ideal sampling rates experimenters can use to extract the best possible data from noise \cite{RN91}.

In Fig. \ref{amp_est}, we show that successful parameter estimation of the underlying OU process depends critically on the known ratio of thermal to multiplicative noise.  From our previous examples with pure thermal and pure multiplicative noise, it seems that if the time series is thermal noise dominated, then the posterior exhibits a clear minimum with the correct parameters. When multiplicative noise dominates, the posterior is ill-formed and drifts into regions of nonphysical parameter space. This behavior is supported by the decrease of the effective sample size (ESS) as the ratio of multiplicative to thermal noise increases (see caption of Fig. \ref{amp_est}).

An unexpected result from our analysis is that adding more thermal noise to the signal will improve the fit under certain conditions. We have shown that HMC fails to distinguish between noise and the signal when the multiplicative noise magnitude exceeds that of the thermal noise. This issue is significant in averaging experiments such as fMRI, where averaging over a particular area of interest may decrease the amount of thermal noise versus multiplicative noise. To address this, it may be possible to add artificial thermal noise to the system, adjusting the noise ratio to a value below one and allowing the removal of multiplicative noise. This could be useful in situations where it was previously impossible to remove multiplicative noise and lead to an increase in the accuracy of further analysis.

\subsection{Appendix}

\subsubsection{Product of two Gaussians}

\begin{equation}
\mathcal{N}(x,\mu_{1},\sigma_{1})\mathcal{N}(x,\mu_{2},\sigma_{2}) 
\propto \mathcal{N}\left(x,\frac{\mu_{1}\sigma_{2}^{2}+\mu_{2}\sigma_{1}^{2}}{\sigma_{1}^2+\sigma_{2}^2},\sqrt{\frac{\sigma_{1}^2\sigma_{2}^2}{\sigma_{1}^2+\sigma_{2}^2}}\right)
\end{equation}

\subsubsection{Convolution of two Gaussians}
\begin{equation}
\int_{-\infty}^{+\infty}\mathcal{N}(x-y,\mu_{1},\sigma_{1})\mathcal{N}(y,\mu_{2},\sigma_{2})dy
	\propto\mathcal{N}\left(x,\mu_{1}+\mu_{2},\sqrt{\sigma_{1}^2+\sigma_{2}^2}\right)
\end{equation}
specifically, to calculate $\alpha(x_{n})$ and $beta(x_{n-1})$ we encounter the following convolutions
\begin{equation}
	\begin{aligned}
	\int_{-\infty}^{+\infty}dx_{n-1}
	\exp \left( -\frac{{{{\left( {x_{n-1} - \mu} \right)}^2}}}{{2\sigma^{2}}} \right)
	\exp \left( -\frac{{{{\left( {x_{n} - Bx_{n-1}} \right)}^2}}}{{2A(1-B^{2})}} \right)
	&\propto \mathcal{N}\left(x_{n},B\mu,\sqrt{B^{2}\sigma^2+A(1-B^{2})}\right)\\
	\int_{-\infty}^{+\infty}dx_{n}
	\exp \left( -\frac{{{{\left( {x_{n} - y_{n}} \right)}^2}}}{{2\sigma_{N}^{2}}} \right)
	\exp \left( -\frac{{{{\left( {x_{n} - Bx_{n-1}} \right)}^2}}}{{2A(1-B^{2})}} \right)
	&\propto \mathcal{N}\left(x_{n-1},y_{n}/B,\sqrt{\sigma^2+A(1-B^{2})}/B\right)
	\end{aligned}
\end{equation}
\subsubsection{Calculate crosscorrelation coefficients}
In Eq. \ref{margxnxnmone} we encounter bivariate distributions such as
\begin{equation}
	\begin{aligned}
	p(x,y)=\frac{1}{S}
	\exp \left( -\frac{{{{\left( {x - \mu_{x}} \right)}^2}}}{{2\sigma_{x}^{2}}} \right)
	\exp \left( -\frac{{{{\left( {y - Bx} \right)}^2}}}{{2A(1-B^{2})}} \right)
	\exp \left( -\frac{{{{\left( {y - \mu_{y}} \right)}^2}}}{{2\sigma_{y}^{2}}} \right)
	\end{aligned}
\end{equation}
where $S$ is the normalization constant
\begin{equation}
	S = \int_{-\infty}^{\infty}\int_{-\infty}^{\infty}p(x,y)dxdy
\end{equation}
resulting in the expectation value of $xy$
\begin{equation}
	\begin{aligned}
	\mathbb{E}(xy) &= \int_{-\infty}^{\infty}\int_{-\infty}^{\infty}p(x,y)xydxdy\\
	&=	\frac{1}{(A(1-B^{2})+B^{2}\sigma^{x}_{x}+\sigma^{2}_{y})^{2}}\\
&\left( A(1-B^{2})(B^{2}\mu_{x}\mu_{y}\sigma_{x}^{2}+B\mu_{y}^{2}\sigma_{x}^{2}+\mu_{x}\mu_{y}\sigma_{y}^{2}+B(\mu_{x}^{2}+\sigma_{x}^{2})\sigma_{y}^{2})\right.\\
	&\left.+B(2B\mu_{x}\mu_{y}\sigma_{x}^{2})\sigma_{y}^{2}+(\mu_{x}^{2}+\sigma_{x}^{2})\sigma_{y}^{2}+B^{2}\sigma_{x}^{2}(\mu_{y}^2+\sigma_{y}^{2}))
	+A^{2}(1-B^{2})^{2}\mu_{x}\mu_{y} \right)
	\end{aligned}
\end{equation}

In this section, we collect useful procedures for calculating the different steps in the EM algorithm.

\subsubsection{OU Second Order Power Spectrum Shape Derivation}

We begin by expanding the definition of the second-order power spectrum for our system with multiplicative noise.

\begin{multline}
E[(X_1 + Y_1)^2(X_2 + Y_2)^2] = E[X_1^2X_2^2 + X_1^2Y_2^2 + 2X_2Y_2X_1^2\\ + Y_1^2X_2^2 + Y_1^2Y_2^2 + 2X_2Y_2Y_1^2 + 2X_1Y_1X_2^2 +2X_1Y_1Y_2^2 + 4X_1X_2Y_1Y_2]
\end{multline}

Because of the symmetry in all the odd-powered terms, the expectation will be zero and will not contribute to the autocorrelation function when $t_1 \neq t_2$. For our purposes, since we are interested in an underlying shape, we can examine the simplified expression, $E[X_1^2X_2^2]$.

We now enforce the assumption that $X_1$ and $X_2$ represent the distribution of an OU process at two different times. It is known that the integral of Brownian motion is equivalent to a zero mean Gaussian with variance (insert citation here), 

\begin{equation}
 \int_{0}^{t} g(s) \,dW_s = N(0,\int_{0}^{t} g(s)^2 \,ds)
\end{equation}

The Ornstein-Uhlenbeck process with zero drift is 
\begin{equation}
x_t = x_0e^{-\theta t} + \sigma \int_{0}^{t} e^{\theta(t-s)} \,dW_s
\end{equation}

For our purposes, we demean, and thus, we can write, 
\begin{equation}
E[x_s^2x_t^2] = \sigma^4e^{-2\theta(s+t)}E[(\int_{0}^{s} e^{\theta(u-s)} \,dW_u)^2(\int_{0}^{t} e^{\theta(v - t)} \,dW_v)^2]
\end{equation}

assuming $s<t$, we can expand the product as

\begin{equation}
x_s^2x_t^2 = \sigma^4e^{-2\theta(s+t)} [ E[(\int_{0}^{s} e^{\theta(u)} \,dW_u)^4] + E[(\int_{0}^{s} e^{\theta(u)} \,dW_u)^2]E[(\int_{s}^{t} e^{\theta(v)} \,dW_v)^2] ]
\end{equation}

we apply Ito-symmetry to the right two terms and use equation 24 for the first term producing,

\begin{equation}
x_s^2x_t^2 = \sigma^4e^{-2\theta(s+t)}[E[N(0, \int_{0}^{s}e^{2\theta u} du)] + \int_{0}^{s}e^{2\theta u} du + \int_{s}^{t}e^{2\theta v} dv
\end{equation}

evaluating the integrals and using the fact that the fourth moment of a Gaussian is $3\sigma^4$ we arrive at the expression 

\begin{equation}
g_x^2(s,t_2) = \frac{\sigma^{4}e^{-2\theta(s+t)}}{2\theta}[\frac{3}{2\theta}e^{4\theta min(s,t)} -2e^{2\theta min(s,t)} + 1 + (e^{2\theta min(s,t)}-1)(e^{2\theta max(s,t)} - e^{2\theta min(s,t)})]
\end{equation}

where we generalised in terms of $min(s,t)$ and $max(s,t)$ as apposed to $s < t$.

\subsubsection{Derivation of the shape of multiplicative noise in a power spectrum}

\begin{equation}
    S(f) = \frac{1}{n} \lvert \sum^{n}_{t=1}e^{-2\pi it \upsilon}x_t \rvert^2
\end{equation}

We can treat the distribution of the power spectrum as the sum of the square of the real and imaginary components

\begin{equation}
    S(f) = \frac{1}{n} \lvert \sum^{n}_{t=1}e^{-2\pi it \upsilon}x_t \rvert^2
\end{equation}

\begin{equation}
    S(f) = \left( \frac{1}{\sqrt{n}}\sum^{n}_{t=1}cos(2\pi t\upsilon)x_t \right)^2 + \left(\frac{1}{\sqrt{n}}\sum^{n}_{t=1}sin(2\pi t\upsilon)x_t \right)^2 = (X_R)^2 + (X_I)^2
\end{equation}

Each term is the sum of weighted Gaussians. In the case of multiplicative noise, these are dependent but not correlated, meaning that the sum is still Gaussian, and we can conclude that $X_I$ and $X_R$ are also Gaussian. Therefore, we solely need to identify the mean and variance. The mean of every random variable is 0 so the mean of the $X_I$ and $X_C$ is also mean zero. We can use the summation property of variances to calculate the variance of these random variables.

\begin{equation}
    \text{Var}(X_R) = \frac{1}{n}\sum^{n}_{t=1}cos^2(2\pi t \upsilon)\sigma_m^2(t)
\end{equation}
\begin{equation}
    \text{Var}(X_I) = \frac{1}{n}\sum^{n}_{t=1}sin^2(2\pi t \upsilon)\sigma_m^2(t)
\end{equation}

where for multiplicative noise, the sigmas are dependent on the time point being used.

We can then describe the shape of $S(f)$ as,

\begin{equation}
    \left( \frac{1}{n}\sum^{n}_{t=1}cos^2(2\pi t \upsilon)\sigma_m^2(t) \right) \chi^2_1 + \left( \frac{1}{n}\sum^{n}_{t=1}sin^2(2\pi t \upsilon)\sigma_m^2(t) \right) \chi^2_1
\end{equation}

Which simplifies to,

\begin{equation}
    \left( \frac{1}{n}\sum^{n}_{t=1}cos^2(2\pi t \upsilon)\sigma_m^2(t) \right) \chi^2_1 + \left( \frac{1}{n}\sum^{n}_{t=1}sin^2(2\pi t \upsilon)\sigma_m^2(t) \right) \chi^2_1
\end{equation}

This can then be further simplified to 

\begin{equation}
    \frac{1}{n}\sum^{n}_{t=1}\sigma^2_m(t)\chi^2_1
\end{equation}

\begin{acknowledgments}
The research presented here was funded by the following:
National Science Foundation/White House Brain Research
Through the Advancing Innovative Technologies (BRAIN) Initiative, United States (NSFNCS-FR 1926781) and the Baszucki
Brain Research Fund, United States. 
\end{acknowledgments}
\bibliography{OUnoise}
\end{document}